%% file: main.tex
\def\year{2019}\relax
\documentclass[letterpaper]{article} 
\usepackage{arXiv}  
\usepackage{times}  
\usepackage{helvet}  
\usepackage{courier}  
\usepackage{url}  
\usepackage{graphicx}  
\usepackage{multirow}
\usepackage{amsmath}
\usepackage{subfig}
\usepackage{multirow}
\usepackage{bbm}

\usepackage{xcolor}
\usepackage{colortbl}
\usepackage{array}
\usepackage{pifont}
\usepackage{booktabs}

\begin{document}
%
\title{Utilizing Class Information for Deep Network Representation Shaping}

\author{Daeyoung Choi\thanks{Authors contributed equally.} and Wonjong Rhee\footnotemark[1] \\
Department of Transdisciplinary Studies\\
Seoul National University\\
Seoul, 08826, South Korea \\
\texttt{\{choid, wrhee\}@snu.ac.kr} 
}
\maketitle
\begin{abstract}
Statistical characteristics of deep network representations, such as sparsity and correlation, are known to be relevant to the performance and interpretability of deep learning. When a statistical characteristic is desired, often an adequate regularizer can be designed and applied during the training phase. Typically, such a regularizer aims to manipulate a statistical characteristic over all classes together. For classification tasks, however, it might be advantageous to enforce the desired characteristic per class such that different classes can be better distinguished. Motivated by the idea, we design two class-wise regularizers that explicitly utilize class information: class-wise Covariance Regularizer (cw-CR) and class-wise Variance Regularizer (cw-VR). cw-CR targets to reduce the covariance of representations calculated from the same class samples for encouraging feature independence. cw-VR is similar, but variance instead of covariance is targeted to improve feature compactness. For the sake of completeness, their counterparts without using class information, Covariance Regularizer (CR) and Variance Regularizer (VR), are considered together. The four regularizers are conceptually simple and computationally very efficient, and the visualization shows that the regularizers indeed perform distinct representation shaping. In terms of classification performance, significant improvements over the baseline and L1/L2 weight regularization methods were found for 21 out of 22 tasks over popular benchmark datasets. In particular, cw-VR achieved the best performance for 13 tasks including ResNet-32/110. 
\end{abstract}


\section{Introduction}

\begin{figure}[t]
\centering
\centerline{\includegraphics[width=8.25cm]{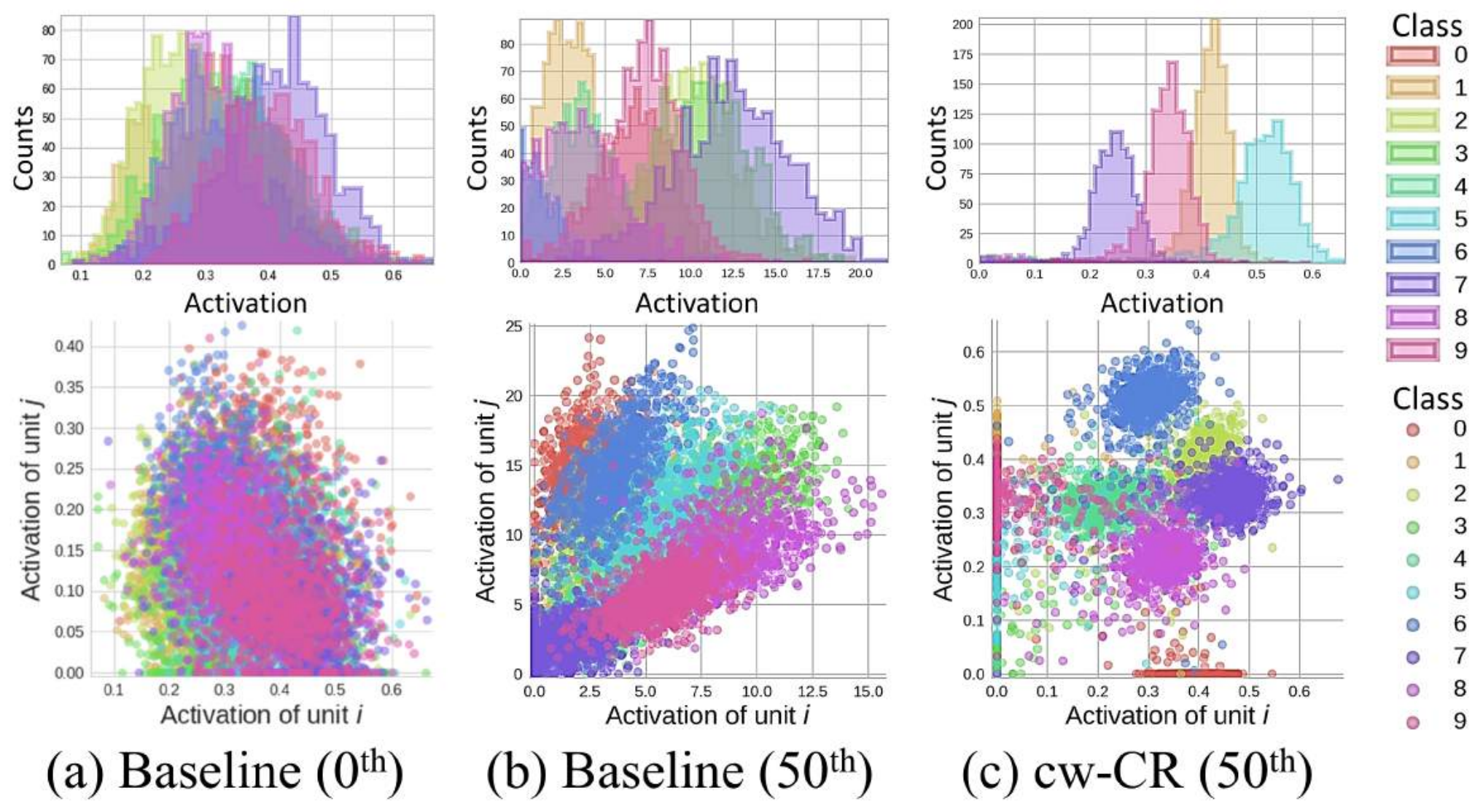}}
\caption{
A single unit's activation histogram (upper three plots) and two randomly chosen units' activation scatter plots (lower three plots) for MNIST. For a 6-layer Multilayer Perceptron (MLP), the fifth layer's representation vectors calculated using 10,000 test samples were used to generate the plots. For the baseline model, a substantial overlap among different classes can be observed at the time of initialization as shown in (a). Even after 50 epochs of training, still, a substantial overlap can be observed as shown in (b). When class information is used to regularize the representation shapes, the overlap is significantly reduced as shown in (c). Note that a slight correlation between each pair of classes can be observed in the scatter plot of (b), but not in that of (c) due to the use of cw-CR. The figures are best viewed in color.
}
\label{fig:mnist_none_hist_scatter}
\end{figure}

For deep learning, a variety of regularization techniques have been developed by focusing on the \textit{weight parameters}. A classic example is the use of L2 \cite{hoerl1970ridge} and L1 \cite{tibshirani1996regression} weight regularizers. They have been popular because they are easy to use, computationally light, and often result in performance enhancements. Another example is the parameter sharing technique that enforces the same weight values as in the Convolutional Neural Networks (CNNs).  
Regularization techniques that focus on the \textit{representation} (the activations of the units in a deep network), however, have been less popular even though the performance of deep learning is known to depend on the learned representation heavily. 

For representation shaping (regularization), some of the promising methods for performance and interpretability include \cite{glorot2011deep,cogswell2015reducing,liao2016learning}.
\cite{glorot2011deep} considers increasing representational sparsity, \cite{cogswell2015reducing} focuses on reducing covariance among hidden units, and \cite{liao2016learning} forces parsimonious representations using k-means style clustering. While all of them are effective representation regularizers, none of them explicitly use class information for the regularization. A few recent works \cite{wen2016discriminative,belharbi2017neural,yang2018robust} do utilize class information, and their approaches are based on \textit{hidden layer activation vectors}. The method of \cite{belharbi2017neural} is computationally expensive because pair-wise dissimilarities need to be calculated among the same class samples in each mini-batch. 

In this work, two computationally light representation regularizers, cw-CR (class-wise Covariance Regularizer) and cw-VR (class-wise Variance Regularizer), that utilize class information are introduced and studied. We came up with the design ideas by observing typical histograms and scatter plots of deep networks as shown in Figure \ref{fig:mnist_none_hist_scatter}. In Figure \ref{fig:mnist_none_hist_scatter} (b), different classes substantially overlap even after the training is complete. If we directly use class information in regularization, as opposed to using it only for cross-entropy cost calculation, we can specifically reduce overlaps or pursue a desired representation characteristic. An example of cw-CR reducing class-wise covariance is shown in Figure \ref{fig:mnist_none_hist_scatter} (c), and later we will show that cw-VR can notably reduce class-wise variance resulting in minimal overlaps. The two class-wise regularizers are very simple and computationally efficient, and therefore can be easily used as L1 or L2 weight regularizers that are very popular. 

\subsection{Our Contributions}
The contributions of this work can be summarized as follows.

\subsubsection{Introduction of three new representation regularizers} 
We introduce two representation regularizers that utilize class information. cw-CR and cw-VR reduce per-class covariance and variance, respectively. In this work, their penalty loss functions are defined, and their gradients are analyzed and interpreted. Also, we investigate VR that is cw-VR's all-class counterpart. Intuitively, reducing the variance of each unit's activations does not make sense unless it is applied per class, but we have tried VR for the sake of completeness and found that VR is useful for performance enhancement. cw-CR's all-class counterpart, CR, is analyzed as well, but CR turns out to be the same as DeCov that was already studied in-depth in \cite{cogswell2015reducing}. 

\subsubsection{Performance improvement with the new representation regularizers}
Rather than trying to find a single case of beating the state-of-the-art record, we performed an extensive set of experiments on the most popular datasets (MNIST, CIFAR-10, CIFAR-100) and architectures (MLP, CNN). Additionally, ResNet \cite{he2016deep} was tested as an example of a sophisticated network, and an image reconstruction task using autoencoder was tested as an example of a different type of task. We have tested a variety of scenarios with different optimizers, number of classes, network size, and data size. The results show that our representation regularizers outperform the baseline (no regularizer) and L1/L2 weight regularizers for almost all the scenarios that we have tested. More importantly, class-wise regularizers (cw-CR, cw-VR) usually outperformed their all-class counterparts (CR, VR). Typically cw-VR was the best performing regularizer and achieved the best performance for the autoencoder task, too.

\subsubsection{Effects of representation regularization}
Through visualizations and quantitative analyses, we show that the new representation regularizers indeed shape representations in the ways that we have intended. The quantitative analysis of representation characteristics, however, indicates that each regularizer affects multiple representation characteristics together and therefore the regularizers cannot be used to control a single representation characteristic without at least mildly affecting some other representation characteristics.


\section{Related Works}

\subsection{Regularization for Deep Learning}
The classic regularizers apply L2 \cite{hoerl1970ridge} and L1 \cite{tibshirani1996regression} 
penalties to the \textit{weights} of models, and they are widely used for Deep Neural Networks (DNNs) as well. 
\cite{wen2016learning} extended L1 regularizers by using group lasso to regularize 
the structures of DNN (i.e., filters, channels, filter shapes, and layer depth).
\cite{srivastava2014dropout} devised dropout that randomly applies activation masking 
over the units.
While dropout is applied in a multiplicative manner, \cite{glorot2011deep} used L1 penalty 
regularization on the activations to encourage sparse representations.
XCov proposed by \cite{cheung2014discovering} minimizes the covariance between 
autoencoding units and label encoding units of the same layer such that 
representations can be disentangled.  
Batch normalization (BN) proposed by \cite{ioffe2015batch} exploits mini-batch statistics 
to normalize activations. It was developed to accelerate training speed by preventing 
internal covariate shift, but it was also found to be a useful regularizer.
In line with batch normalization, weight normalization, developed by \cite{salimans2016weight}, 
uses mini-batch statistics to normalize weight vectors. 
Layer normalization proposed by \cite{ba2016layer} is a RNN version of batch normalization,
where they compute the mean and variance used for normalization from all of the summed
inputs to the units in a layer on a single training case.
There are many other publications on regularization techniques for deep learning,
but we still do not fully understand how they really affect the performance.  
Recent work by \cite{zhang2016understanding}
shows that the traditional concept of controlling generalization error by regularizing the effective capacity does not apply to the modern DNNs.

\subsection{Penalty Regularization on Representations}
Some of the existing regularization methods explicitly shape representations by adopting a penalty regularization term.
DeCov \cite{cogswell2015reducing} is a penalty regularizer that minimizes the off-diagonals of a layer's representation covariance matrix. DeCov reduces co-adaptation of a layer's units by encouraging the units to be decorrelated. In this work, 
it is called as CR (Covariance Regularizer) for consistent naming.
A recent work \cite{liao2016learning} used a clustering based regularization that encourages parsimonious representations. In their work, similar representations in sample, spatial, and channel dimensions are clustered and used for regularization such that similar representations are encouraged to become even more similar. While their work can be applied to unsupervised as well as supervised tasks, our work utilizes a much simpler and computationally efficient method of directly using class labels during training to avoid k-means like clustering. 

\subsection{Class-wise Learning}
True class information has been rarely used directly for regularization methods.
Traditionally, the class information has been used only for evaluating the correctness of
predictions and the relevant cost function terms. Some of the recent works, however, 
have adopted the class-wise concept in more sophisticated ways. In those works, 
class information is used as a switch or for emphasizing the discriminative aspects over different classes. 
As an example, \cite{li2008kernel} proposed a kernel learning method using class information to model the manifold structure. They modify locality preserving projection to be class dependent. \cite{jiang2011learning} 
added label consistent regularizers for learning a discriminative dictionary. 
\cite{wen2016discriminative} developed a regularizer called center loss that reduces the activation vector distance between representations and their corresponding class centers for face recognition tasks.
\cite{yang2018robust} designed a loss function named prototype loss that improves representation's intra-class compactness for enhancing the robustness of CNN.
Another recent work by \cite{belharbi2017neural} directly uses class labels to encourage similar representations per class as in our work, but it is computationally heavy as explained earlier.  
Besides the pair-wise computation, two optimizers are used for handling the supervised loss term and the hint term separately. 
Class information is used for autoencoder tasks as well. \cite{shi2016learning} implicitly reduced the intra-class variation of reconstructed samples by minimizing pair-wise distances among same class samples.
Like the strategies listed above, our cw-VR and cw-CR use class-wise information to control the statistical characteristics of representations. However, our methods are simple because only one optimizer is used, and computationally efficient because pair-wise computation is not required.


\section{Class-wise Representation Regularizers: cw-CR and cw-VR}

In this section, we first present basic statistics of representations. Then, three representation regularizers, cw-CR, cw-VR, and VR are introduced with their penalty loss functions and gradients. Interpretations of the loss functions and gradients are provided as well. 

\subsection{Basic Statistics of Representations}
\label{subsection:stats}
For the layer $l$, the output activation vector of the layer is defined as 
$\mathbf{z}_l = \max(\mathbf{W}^\top_l \mathbf{z}_{l-1} + \mathbf{b}_l, 0)$ using Rectified Linear Unit (ReLU)
activation function. Because we will be focusing on the layer $l$ for most of the explanations, 
we drop the layer index. 
Then, $z_i$ is the $i^{th}$ element of $\mathbf{z}$ (i.e. activation of $i^{th}$ unit). 

To use statistical properties of representations, we define mean of unit $i$, $\mu_i$, and covariance 
between unit $i$ and unit $j$, $\textit{c}_{i,j}$, using the $N$ samples in each mini-batch. 
\begin{align}
    \mu_i &= \frac{1}{N} \sum_n z_{i,n}  \label{eq:mean}  \\
    \textit{c}_{i,j} &= \frac{1}{N} \sum_n (z_{i,n} - \mu_i)(z_{j,n} - \mu_j) \label{eq:covariance}
\end{align}
Here, $z_{i,n}$ is the activation of unit $i$ for $n^{th}$ sample in the mini-batch.  
From equation (\ref{eq:covariance}), variance of $i$ unit can be written as the following. 
\begin{align}
    \textit{v}_{i} &= \textit{c}_{i,i} \label{eq:variance}
\end{align}
When class-wise statistics need to be considered, we choose a single label $k$ from $K$ labels
and evaluate mean, covariance, and variance using only the data samples with true label $k$
in the mini-batch. 
\begin{align}
    \mu_i^k &= \frac{1}{|S_k|} \sum_{n \in S_k} z_{i,n} \label{eq:mean_cw} \\
    \textit{c}_{i,j}^k &= \frac{1}{|S_k|} \sum_{n \in S_k} (z_{i,n} - \mu_i^k)(z_{j,n} - \mu_j^k) \label{eq:covariance_cw}  \\  
    \textit{v}_{i}^k &= \textit{c}_{i,i}^k   \label{eq:variance_cw}
\end{align}
Here, $S_k$ is the set containing indexes of the samples whose true label is $k$, 
and $|S_k|$ is the cardinality of the set $S_k$.

\begin{table*}[t]
\caption{Penalty loss functions and gradients of the representation regularizers. All the penalty loss functions are normalized with the number of units ($I$) and the number of classes ($K$) such that the value of $\lambda$ can have a consistent meaning. CR and cw-CR are standardized using the number of distinct covariance combinations.}
\centering
\begin{tabular}{rlrl}
		\hline
		\multicolumn{2}{c}{Penalty loss function}  & \multicolumn{2}{c}{Gradient}  \\ \hline
		$\displaystyle{\Omega}_{CR}$ & $\displaystyle=\frac{2}{I(I-1)}\sum_{i\neq j} (c_{i,j})^{2} $    & $\displaystyle\frac{\partial{{\Omega}_{CR}}}{\partial{z_{i,n}}}$ & $\displaystyle=\frac{4}{NI(I-1)}\sum_{j\neq{i}}^{}{c_{i,j}(z_{j,n}-\mu_{j}})$  \\ 
		
		$\displaystyle{\Omega}_{cw{\text -}CR}$ & $\displaystyle=\frac{2}{KI(I-1)}\sum_k \sum_{i\neq j} (c_{i,j}^{k})^{2} $   & $\displaystyle\frac{\partial{{\Omega}_{cw{\text-}CR}}}{\partial{z_{i,n}}}$ & $\displaystyle=\frac{4}{KI(I-1)|S_k|}\sum_{j\neq{i}}^{}{c_{i,j}^{k}(z_{j,n}-\mu_{j}^{k}}),  n \in S_k$  \\ 
		
		$\displaystyle{\Omega}_{VR}$ & $\displaystyle=\frac{1}{I}\sum_i v_{i}$                                               & $\displaystyle\frac{\partial{{\Omega}_{VR}}}{\partial{z_{i,n}}}$ & $\displaystyle=\frac{2}{NI}(z_{i,n}-\mu_{i})$  \\
		
		$\displaystyle{\Omega}_{cw{\text -}VR}$ & $\displaystyle=\frac{1}{K I}\sum_k \sum_i v_{i}^k $                        & $\displaystyle\frac{\partial{{\Omega}_{cw{\text -}VR}}}{\partial{{z}_{i,n}}}$ & $\displaystyle =\frac{2}{KI|S_k|}({z}_{i,n}-{\mu}_{i}^{k}), n \in S_k$  \\ \hline
	\end{tabular}
\label{table:loss_function}
\end{table*}

\subsection{cw-CR}
cw-CR uses off-diagonal terms of the mini-batch covariance matrix of activations per class as the penalty term: ${\Omega}_{cw{\text -}CR}=\sum_k \sum_{i\neq j} (c_{i,j}^{k})^{2}$. This term is added to the original cost function $J$, and the total cost function $\widetilde{J}$ can be denoted as  
\begin{align}
    \widetilde{J}=J+\lambda{\Omega}_{cw{\text -}CR}(\mathbf{z}),
\end{align}
where $\lambda$ is the penalty loss weight ($\lambda \in [0, \infty)$). The penalty loss weight balances between the original cost function $J$ and the penalty loss term $\Omega$. When $\lambda$ is equal to zero, $\widetilde{J}$ is the same as $J$, and cw-CR does not influence the network. When $\lambda$ is a positive number, the network is regularized by cw-CR, and the performance is affected. In practice, we have observed that deep networks with too large $\lambda$ cannot be trained at all.

\subsection{cw-VR}
A very intuitive way of enforcing distinguished representations per class is to maximize the inter-class distances in the representation space. 
Because inter-class needs to be maximized, the corresponding penalty term can be inverted or multiplied by -1 before it is minimized with the original cost function. 
We tried such approaches, but the optimization became unstable (failed to converge).
An alternative way is to reduce intra-class (same-class) variance. By applying this idea, the penalty loss term of cw-VR can be formulated as ${\Omega}_{cw{\text -}VR}=\sum_k \sum_i v_{i}^k$. 

With the design of cw-VR, we naturally invented VR that is the all-class counterpart of cw-VR. VR minimizes the activation variance of each unit, and it is mostly the same as cw-VR except for not using the class information. We expected VR to hurt the performance of deep networks because it encourages all classes to have similar representation in each unit. VR, however, turned out to be effective and useful for performance enhancement. We provide a possible explanation in the Experiments section. 

\subsection{Penalty Loss Functions and Gradients}

The penalty loss functions of cw-CR and cw-VR are similar to CR and VR, respectively, except that the values are calculated for each class using the mini-batch samples with the same class label. Also, gradients of CR and cw-CR are related to those of VR and cw-VR as shown in Table \ref{table:loss_function}. We investigate more details of the equations in the following.

\subsubsection{Interpretation of the gradients}
Among the gradient equations shown in Table \ref{table:loss_function}, the easiest to understand is VR's gradient. It contains the term ${z}_{i,n}-{\mu}_{i}$, indicating that the representation ${z}_{i,n}$ of each sample $n$ is encouraged to become closer to the mean activation ${\mu}_{i}$. In this way, each unit's variance can be reduced. For cw-VR, the equation contains ${z}_{i,n}-{\mu}_{i}^{k}$ instead of ${z}_{i,n}-{\mu}_{i}$. Therefore the representation ${z}_{i,n}$ of a class $k$ sample is encouraged to become closer to the \textit{class} mean activation ${\mu}_{i}^{k}$. Clearly, the variance reduction is applied per class by cw-VR. 

For CR, the equation is less straightforward. As explained in \cite{cogswell2015reducing}, a possible interpretation is that the covariance term $c_{i,j}$ is encouraged to be reduced where $z_{j,n}-\mu_j$ acts as the weight. But, another possible interpretation is that $z_{j,n}$ is encouraged to become closer to $\mu_j$ just as in the case of VR, where $c_{i,j}$ acts as the weight. Note that VR's mechanism is straightforward where each unit's variance is directly addressed in the gradient equation of activation $i$, but CR's mechanism is slightly complicated where all variances over all activations of $j$ ($j=1,...,I$, where $j \neq i$) are collectively addressed through the summation terms over all $j$ ($j=1,...,I$, where $j \neq i$). Thus, one can interpret CR as a hybrid regularizer that wants either or both of covariance and variance to be reduced. This can be the reason why the visualizations of CR and VR are similar as will be shown in Figure \ref{fig:representation} later. 

For cw-CR, it can be interpreted similarly. As in the relationship between VR and cw-VR, cw-CR is the class-wise counterpart of CR and it can be confirmed in the gradient equation: cw-CR has $c_{i,j}^k({z}_{j,n}-{\mu}_{j}^{k})$ instead of $c_{i,j}({z}_{j,n}-{\mu}_{j})$. As in our explanation of CR, cw-CR can also be interpreted as trying to reduce either or both of covariance and variance. The visualizations of cw-CR and cw-VR turn out to be similar as well. 

The interpretations can be summarized as follows. VR and cw-VR aim to reduce activation variance whereas CR and cw-CR additionally aim to reduce covariance. CR and VR do not distinguish among different classes, but cw-CR and cw-VR explicitly perform representation shaping per class.

\subsubsection{Activation squashing effect}
There is another important effect that is not necessarily obvious from the gradient formulations.
For L1W (L1 weight regularization) and L2W (L2 weight regularization), the gradients contain the weight terms, and therefore the weights are explicitly encouraged to become smaller. Similarly, our representation regularizers include the activation terms $z_{i,n}$ and therefore the activations are explicitly encouraged to become smaller (when activations become close to zero, the mean terms become close to zero as well). Thus, a simple way to reduce the penalty loss is to scale the activations to small values instead of satisfying the balance between the terms in the gradient equations. 
This means that there is a chance for the learning algorithm to squash activations just so that the representation regularization term can be ignored. As we will see later in the next section, indeed activation squashing happens when our regularizers are applied. Nonetheless, we will also show that the desired statistical properties are sufficiently manifested anyway. One might be able to prevent activation squashing with another regularization technique, but such an experiment was not in the scope of this work.


\section{Experiments}
\label{sec:experiments}
In this section, we investigate performance improvements of the four representation regularizers, where baseline, L1W, L2W, CR, cw-CR, VR, and cw-VR are evaluated for image classification and reconstruction tasks. When a regularizer (including L1W and L2W) was used for an evaluation scenario, the penalty loss weight $\lambda$ was determined as one of \{0.001, 0.01, 0.1, 1, 10, 100\} using 10,000 validation samples. Once the $\lambda$ was determined, performance evaluation was repeated five times. Code is made available at
\url{https://github.com/snu-adsl/class_wise_regularizer}.

\subsection{Image Classification Task}
\begin{table}[t]
\centering
\caption{Error performance (\%) for CIFAR-10 CNN model.}
\label{table:cifar-10}
\begin{tabular}{ccc}
\hline
\multirow{2}{*}{Regularizer} & \multicolumn{2}{c}{Optimizer}                      \\ \cline{2-3} 
                             & Adam                                    & Momentum \\ \hline
Baseline                     & $26.64 \pm 0.16$                        & $25.78 \pm 0.37$ \\ \hline
L1W                          & $26.46 \pm 0.39$                        & $25.73 \pm 0.40$ \\
L2W                          & $25.71 \pm 0.98$                        & $26.35 \pm 0.54$ \\ \hline
CR                           & $24.96 \pm 0.63$                        & $26.72 \pm 0.61$ \\ 
cw-CR                        & $22.99 \pm 0.58$                        & $25.93 \pm 0.59$ \\
VR                           & \pmb{$21.44 \pm 0.88$}                  & $25.01 \pm 0.41$ \\
cw-VR                        & $21.58 \pm 0.21$                        & \pmb{$24.42 \pm 0.31$} \\ \hline
\end{tabular}
\end{table}
\begin{table}[t]
\centering
\caption{Error performance (\%) for CIFAR-100 CNN model.}
\label{table:cifar-100}
\resizebox{\columnwidth}{!}{%
\begin{tabular}{cccc}
\hline
\multirow{2}{*}{Regularizer} & \multicolumn{3}{c}{Number of Classes}                                                                                       \\ \cline{2-4}
                                                                   & 16                                      & 64            & 100                         \\ \hline
Baseline    & $45.75 \pm 0.73$                  & $58.02 \pm 0.40$       & $61.26 \pm 0.52$                 \\ \hline
L1W         & $45.08 \pm 1.53$                  & $58.08 \pm 1.18$       & $60.97 \pm 0.64$                 \\
L2W         & $45.28 \pm 1.59$                  & $57.47 \pm 0.66$       & $60.23 \pm 0.31$                 \\ \hline
CR          & $44.55 \pm 1.10$                  & $56.76 \pm 0.86$       & $59.88 \pm 0.50$                 \\ 
cw-CR       & $43.50 \pm 1.21$                  & $54.24 \pm 0.64$       & $57.03 \pm 0.73$                 \\
VR          & $42.33 \pm 1.03$                  & $54.32 \pm 0.40$       & $57.68 \pm 0.94$                  \\
cw-VR       & \pmb{$41.38 \pm 0.53$}            & \pmb{$54.23 \pm 1.06$} & \pmb{$56.75 \pm 0.64$}             \\ \hline
\end{tabular}
}
\end{table}

Three popular datasets (MNIST, CIFAR-10, and CIFAR-100) were used as benchmarks. An MLP model was used for MNIST, and a CNN model was used for CIFAR-10/100. The details of the architecture hyperparameters can be found in Section A of the supplementary materials. All the regularizers were applied to the fifth layer of the 6-layer MLP model and the fully connected layer of the CNN model, and the reason will be explained in the Layer Dependency section. For L1W and L2W, we applied regularization to all the layers as well for comparison, but the performance results were comparable to when applied to the fifth layer. Mini-batch size was increased to 500 for CIFAR-100 such that class-wise operations can be appropriately performed but was kept at the default value of 100 for MNIST and CIFAR-10. We have tested a total of 20 scenarios where the choice of an optimizer, number of classes, network size, or data size was varied.

The results for two CIFAR-10 CNN scenarios are shown in Table \ref{table:cifar-10} and three CIFAR-100 CNN scenarios are shown in Table \ref{table:cifar-100}. The rest of the scenarios including full cases of MNIST MLP can be found in Section B of the supplementary materials. In the Table \ref{table:cifar-10} and Table \ref{table:cifar-100}, it can be seen that cw-VR achieves the best performance in 4 out of 5 cases and class-wise regularizers perform better than their all-class counterparts except for one case. For the scenarios shown in Table \ref{table:cifar-100}, we initially guessed that the performance of class-wise regularizers would be sensitive to the number of classes, but cw-VR performed well for all three cases. As for the 20 scenarios that were tested, the best performing one was cw-VR for 11 cases, VR for 5 cases, cw-CR for 2 cases, and CR for 1 case. L1W and L2W were never the best performing one, and the baseline (no regularization) performed the best for only one case. 

As mentioned earlier, in general, VR did not hurt performance compared to the baseline. There are two possible explanations. First, representation characteristics other than variance are affected together by VR (see Table \ref{table:statistical_property} in the next section), and VR might have indirectly created a positive effect. Second, the cross-entropy term limits how much VR performs variance reduction, and the overall effects might be more complicated than a simple variance reduction.


To test a sophisticated and advanced DNN architecture, we tried the four representation regularizers on ResNet-32/110. ResNet is known as one of the best performing deep networks for CIFAR-10, and we applied the four representation regularizers to the output layer without modifying the network's architecture or hyperparameters. The results are shown in Table \ref{table:resnet-110}. All four turned out to have positive effects where cw-VR showed the best performance again. 

\begin{table}[t]
\centering
\caption{Error performance (\%) for ResNet-32/110 (CIFAR-10). 
For ResNet-32, average of two experiments is shown. For ResNet-110,
we experimented five times and \lq best (mean$\pm$std)\rq \ is reported as in \cite{he2016deep}.
}
\resizebox{\columnwidth}{!}{%
\begin{tabular}{lcc}
\hline
\multicolumn{1}{c}{Model \& Regularizer}   & He et al. & Ours      \\ 
\hline
ResNet-32                      & 7.51     & 7.39  \\       
ResNet-32 + CR                 &                   & 7.27           \\ 
ResNet-32 + cw-CR              &                   & 7.21           \\
ResNet-32 + VR                 &                   & 7.22           \\
ResNet-32 + cw-VR              &                   & \textbf{7.17}  \\ \hline
ResNet-110                      & 6.43 \small{(6.61$\pm$0.16)}     & 6.12 \small{(6.31$\pm$0.14)}  \\  
ResNet-110 + CR                 &                   & 6.17 \small{(6.26$\pm$0.05)}           \\ 
ResNet-110 + cw-CR              &                   & 6.10 \small{(6.18$\pm$0.10)}           \\
ResNet-110 + VR                 &                   & 6.10 \small{(6.17$\pm$0.05)}           \\
ResNet-110 + cw-VR              &                   & \textbf{6.00} \small{(6.18$\pm$0.15)}  \\ \hline
\end{tabular}
}
\label{table:resnet-110}
\end{table}

\begin{figure*}[t]
\centering
\centerline{\includegraphics[width=1\textwidth]{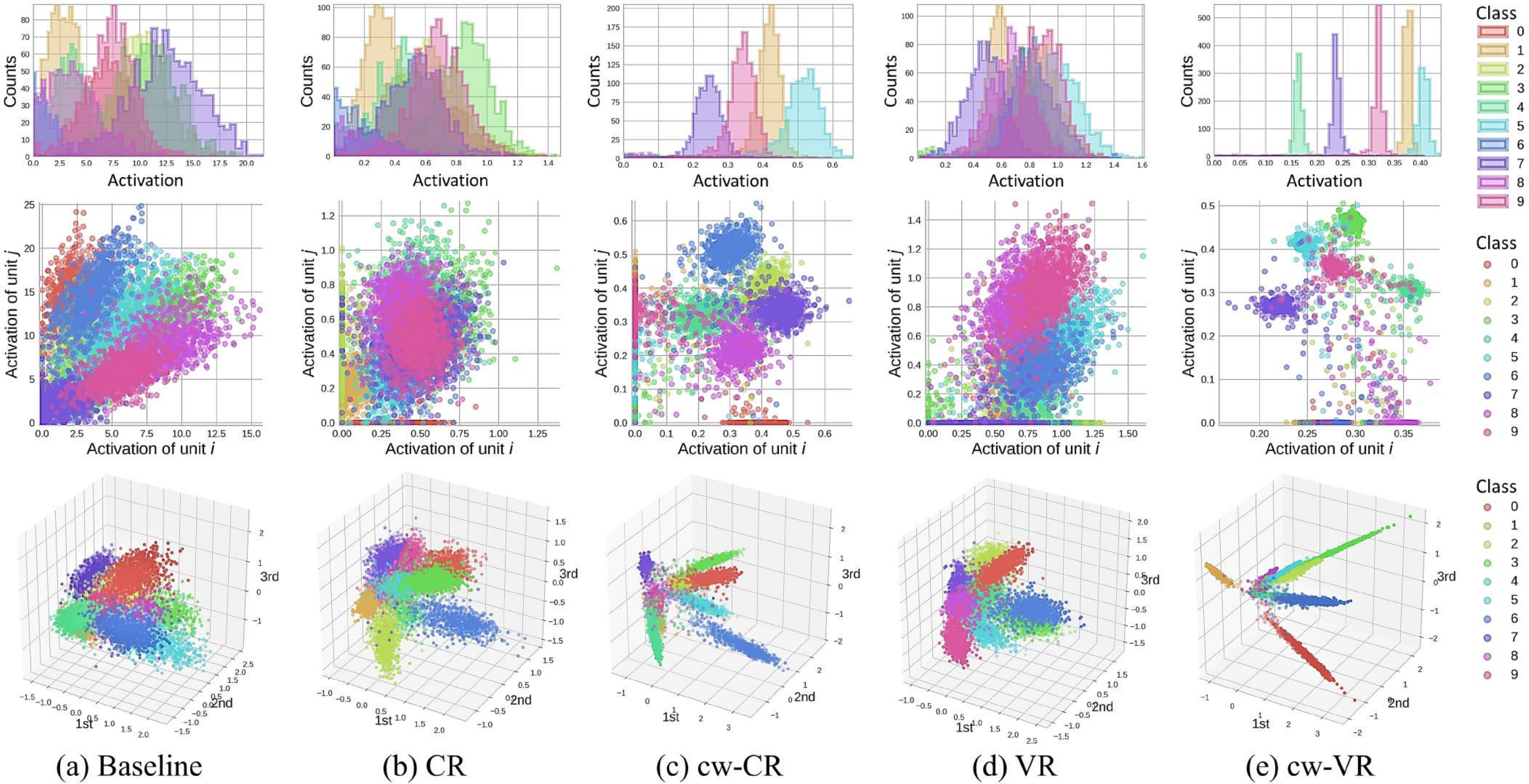}}
\caption{Visualization of the learned representations for MNIST. The plots in top and middle rows were generated in the same way as in the Figure \ref{fig:mnist_none_hist_scatter}. The plots in the bottom row show the top three principle components of the representations. 
}
\label{fig:representation}
\end{figure*}

\begin{table*}[t]
\centering
\caption{Quantitative evaluations of representation characteristics. 
}
\label{table:statistical_property}
\resizebox{\textwidth}{!}{%
\begin{tabular}{cccccccc}
\hline
Regularizer &	 Test error (\%)        &    	 \textsc{Activation\_amplitude}  &	 \shortstack{\textsc{Covariance} \\ (CR)}   &	 \shortstack{\textsc{Correlation} \\ (CR)} & 	 \shortstack{\textsc{cw\_Correlation} \\ (cw-CR)}   &	 \shortstack{\textsc{Variance} \\ (VR)}  &	 \shortstack{\textsc{N\_cw\_Variance} \\ (cw-VR)} \\ \hline
Baseline    &	 $2.85 \pm 0.11$   &     	 4.93             &    	2.08 &	 0.27               &    	 0.21               &        	 9.05             &  	 1.33                     \\ \hline
L1W        & 	 $2.85 \pm 0.06$   &     	 4.53              &   	1.95 &	 0.28               &    	 0.22               &         	 7.78             &  	 1.33                     \\
L2W        & 	 $3.02 \pm 0.40$   &     	 4.76             &    	2.23 &	 0.29               &    	 0.21                &       	 8.38             &  	 1.36                     \\ \hline
CR          &	 $2.50 \pm 0.05$   &    	 \textit{0.50}    &    	0.01 &	 \textbf{0.19}      &    	 0.15                 &       	 0.04             &  	 1.37                     \\
cw-CR      & 	 $2.49 \pm 0.10$   &     	 \textit{0.63}     &   	0.02 &	 0.31              &    	 \textbf{0.19}        &       	 0.06             &  	 0.95                     \\
VR        &  	 $2.65 \pm 0.11$   &     	 \textit{1.35}     &   	0.15 &	 0.26               &   	 0.17                 &       	 \textbf{0.58}    &  	 1.52                     \\
cw-VR      & 	 \pmb{$2.42 \pm 0.06$}  &	 \textit{0.63}     &   	0.02 &	 0.36               &    	 0.25                 &     	 0.05             &  	 \textbf{0.74}            \\ \hline

\end{tabular}
}
\end{table*}

\subsection{Image Reconstruction Task}
In order to test a completely different type of task, we examined an image reconstruction task where a deep autoencoder are used. Class information is used for representation regularization only. A 6-hidden layer autoencoder with a standard L2 objective function was used. Representation regularizers were only applied to the third layer because the representations of the layer are considered as latent variables. The other experiment settings are the same as the image classification tasks in the previous subsection. The reconstruction error of the baseline is $1.44 \times 10^{-2}$ and become reduced to $1.19 \times 10^{-2}$ when cw-VR is applied. Result details can be found in Section B of the supplementary materials.
As in the classification tasks, class-wise regularizers performed better than their all-class counterparts.


\section{Representation Characteristics}

In this section, we investigate representation characteristics when the regularizers are applied. 

\subsection{Visualization}
In Figure \ref{fig:representation}, the $50^{th}$ epoch plots are shown for the baseline and four representation regularizers. L1W and L2W are excluded because their plots are very similar to those of the baseline.
Principle Component Analysis (PCA) was also performed over the learned representations, and the plots in the bottom row show the top three principal components of the representations (before ReLU).
The first thing that can be noticed is that the representation characteristics are quite different depending on which regularizer is used. Apparently, the regularizers are effective at affecting representation characteristics. 
In the first row, it can be seen that cw-VR minimizes the activation overlaps among different classes as intended. Because the gradient equation of cw-CR is related to that of cw-VR, cw-CR also shows reduced overlaps. CR and VR still show substantial overlaps because class information was not used by them. 
In the second row, a linear correlation can be observed in the scatter plot of the baseline, but such a linear correlation is mostly removed for CR as expected. For VR, still, linear correlations can be observed. For cw-CR and cw-VR, it is difficult to judge because many points do not belong to the main clusters and their effects on correlation are difficult to guess. As we will see in the following quantitative analysis section, in fact, correlation was not reduced for cw-CR and cw-VR.
In the third row, it can be seen that the cw-VR has the least overlaps when the first three principal components are considered. Interestingly, a needle-like shape can be observed for each class in the cw-VR's plot. The plots using learned representations after ReLU are included in Section C of the supplementary materials. Overall, cw-VR shows the most distinct shapes compared to the baseline. 

\subsection{Quantitative Analysis}
For the same MNIST task that was used to plot Figure \ref{fig:mnist_none_hist_scatter} and Figure \ref{fig:representation}, the quantitative values of representation characteristics were evaluated, and the results are shown in Table \ref{table:statistical_property}. Each is calculated using only positive activations and is the average of representation statistics. For example, \textsc{Activation\_amplitude} is the mean of positive activations in a layer.
In the third column (\textsc{Activation\_amplitude}), it can be confirmed that indeed the four representation regularizers cause activation squashing. Nonetheless, the error performance is improved as shown in the second column. For CR, covariance is supposed to be reduced. In the fourth column (\textsc{Covariance}), it can be confirmed that the covariance of CR is much smaller than that of the baseline. The small value, however, is mostly due to the activation squashing. In the fifth column (\textsc{Correlation}), the normalized version of covariance is shown. The correlation of CR is confirmed to be smaller than that of the baseline, but the reduction rate is much smaller compared to the covariance that was affected by the activation squashing. In any case, CR indeed reduces correlation among hidden units. For cw-CR, class-wise correlation (\textsc{cw\_Correlation}) is expected to be small, and it is confirmed in the sixth column. The value 0.19, however, is larger than CR's 0.15 or VR's 0.17. This is an example where not only cw-CR but also other representation regularizers end up reducing \textsc{cw\_Correlation} because the regularizers' gradient equations are related. For VR, variance should be reduced. In the seventh column (\textsc{Variance}), the variance of VR is indeed much smaller than that of the baseline, but again other representation regularizers have even smaller values because their activation squashing is more severe than that of VR. For cw-VR, class-wise variance is supposed to be small. Normalized class-wise variance is shown in the last column (\textsc{N\_cw\_Variance}), and it is confirmed that cw-VR is capable of reducing \textsc{N\_cw\_Variance}. (Normalization was performed by mapping activation range of each hidden unit to [0,10] such that activation squashing effect can be removed.)


\section{Layer Dependency}
\begin{figure}[t]
    \centering
    \subfloat[MNIST]{{\includegraphics[width=4.1cm]{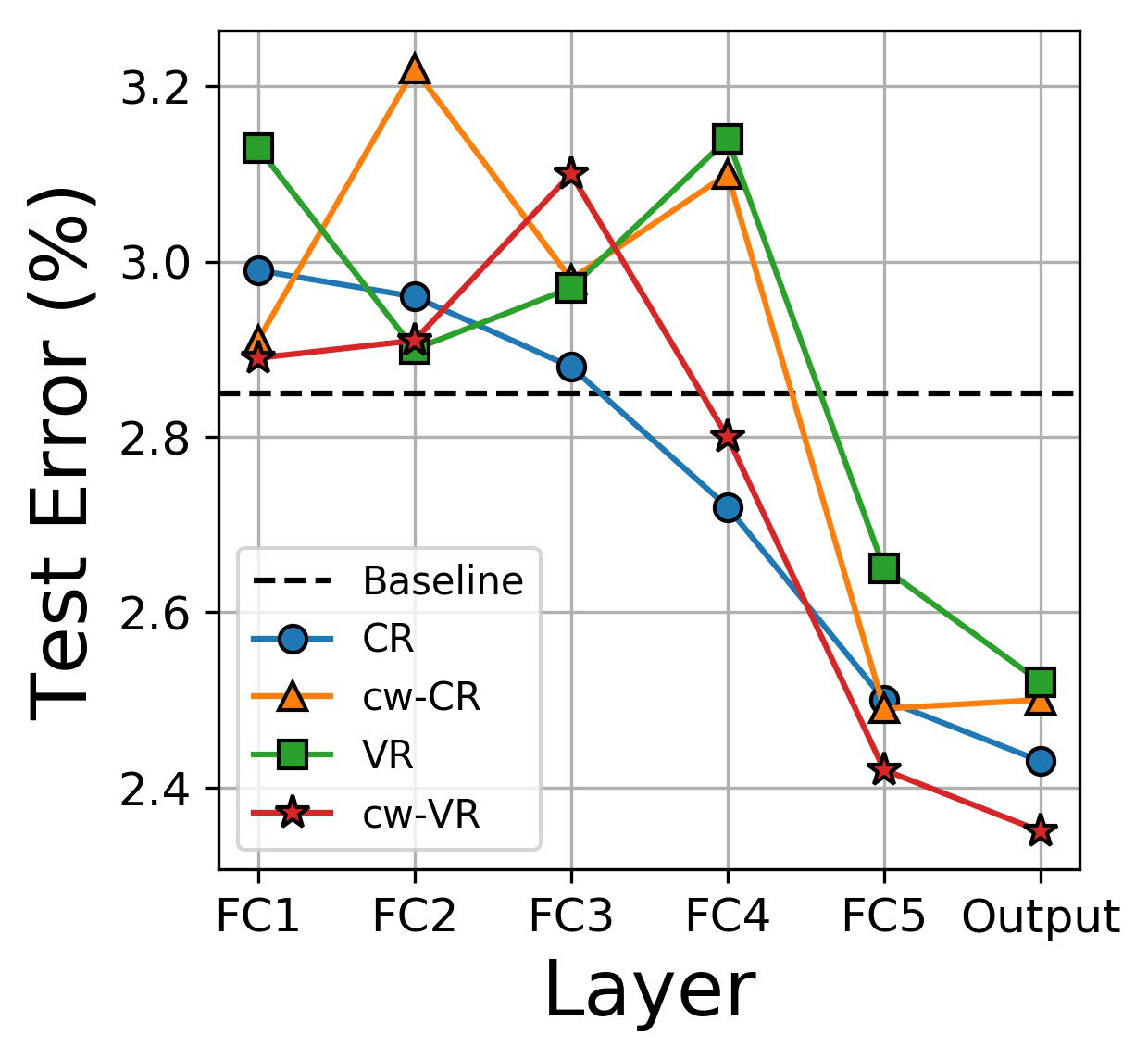} }}%
    \hspace{-0.4\baselineskip}
    \subfloat[CIFAR-100]{{\includegraphics[width=4.1cm]{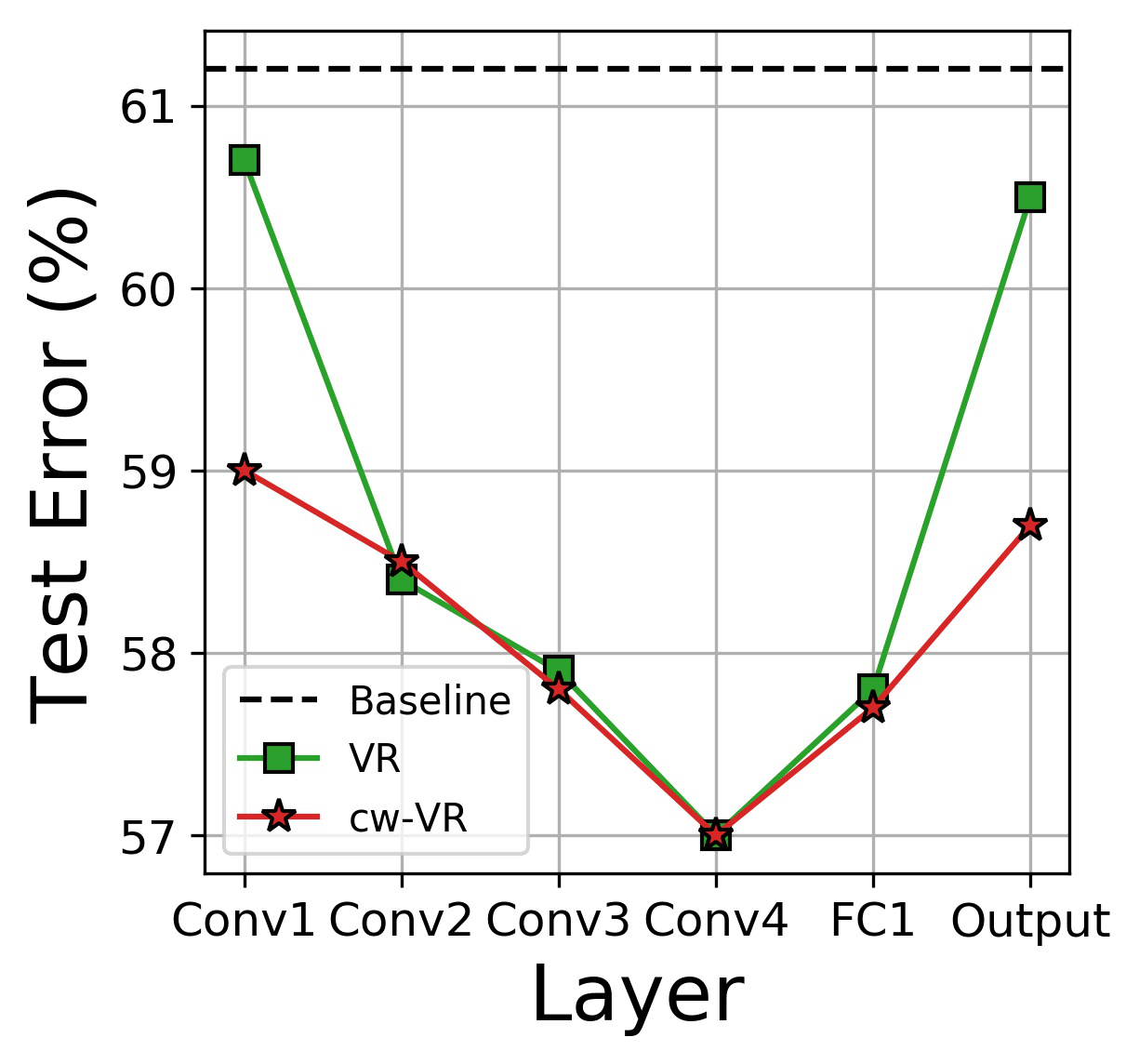} }}%
    \caption{Layer dependency of representation regularizers. The x-axis indicates layers where regularizers are applied. CR and cw-CR are excluded in (b) due to the high computational burden of applying them to the convolutional layers. The result of CIFAR-10 can be found in Section D of the supplementary materials.}%
    \label{fig:layer_dependency}%
\end{figure}

In the previous sections, we have consistently applied the representation regularizers to the upper layers that are closer to the output layer. This is because we have found that it is better to target the upper layers, and two exemplary results are shown in Figure \ref{fig:layer_dependency}. In Figure \ref{fig:layer_dependency} (a), the performance improvement becomes larger as the representation regularization targets upper layers. In fact, the best performance is observed when the output layer is regularized. In Figure \ref{fig:layer_dependency} (b), similar patterns can be seen over the convolutional layers, but the performance degrades when applied to fully connected or output layers. This phenomenon is probably relevant to how representations are developed in deep networks. Because the lower layers often represent many simpler concepts, regularizing the shapes of representations can be harmful. For the upper layers, a smaller number of more complex concepts are represented and therefore controlling representation characteristics (e.g., reduction of activation overlaps) might have a better chance to improve the performance. 

\section{Discussion and Conclusion}
A well-known representation regularizer is L1 representation regularizer (L1R) whose penalty loss function can be written as ${\Omega}_{L1R}=\frac{1}{NI}\sum_n \sum_i |z_{i,n}|$. L1R is known to increase representational sparsity. CR and VR have second-order terms in their penalty loss functions, but L1R does not. As a consequence, L1R's class-wise counterpart turns out to have the same penalty function as L1R's (this is trivial to prove). So, one might say that L1R is also a class-wise representation regularizer just like cw-CR and cw-VR. When it is used, however, there is no need for the true class information. For instance, when true label information is not available for an autoencoder problem, one might use L1R and still have a chance to obtain the benefits of class-wise regularization. In our study, we have not included L1R such that we can better focus on the difference between all-class and class-wise regularizers. When cw-VR was directly compared with L1R in terms of performance, we have found that cw-VR performs better than L1R for 12 out of the 21 test scenarios (ResNet-110 and an autoencoder were not tested). Overall, however, it looks like both L1R and cw-VR are very effective representation regularizers for improving performance of deep networks. 

Dropout and batch normalization are very popular regularizers, but they are fundamentally different because they are not \lq penalty cost function' regularizers. Instead, they are implemented by directly affecting the feedforward calculations during training. Dropout has been shown to have similar effects as ensemble and data-augmentation through its noisy training procedure, and such benefits are not obtainable with a penalty regularizer. On the other hand, there is a common belief that \lq dropout reduces co-adaptation (or pair-wise correlation).\rq \,Reducing correlation is something that can be done by penalty regularizers as we have shown in this work. When we applied the same quantitative analysis on the test scenarios while using dropout, however, we have found that dropout does not really reduce the correlation. This indicates that the belief might be an incorrect myth. 
Batch normalization has been known to have a stabilization effect because it can adjust covariate shift even when the network is in the early stage of training. Thus a higher learning rate can be used for faster training. Such an effect is not something that can be achieved with a penalty regularizer. But when dropout and batch normalization were directly compared with the two representation regularizers cw-VR and L1R in terms of performance, we have found that at least one of cw-VR and L1R outperforms both of dropout and batch normalization for 16 out of the 20 test cases (ResNet-32/110 and an autoencoder were not tested).
Despite the performance results for our benchmark scenarios, it is important to recognize that dropout and batch normalization might be able to play completely different roles that cannot be addressed by the penalty regularizers. When such additional roles are not important for a task as in our test scenarios, there is a very high chance of penalty regularizers outperforming dropout and batch normalization.

Performance improvement through representation regularizers, especially by utilizing class information, has been addressed in this work and other previous works. The underlying mechanism for the improvement, however, is still unclear. Recently, \cite{choi2018statistical} showed that
some of the statistical properties of representations cannot be the direct cause of performance improvement. The representation regularizers might have tuning effects instead. 

With the enormous efforts of the research community, deep learning is becoming better understood, and regularization techniques are evolving with the in-depth understandings. In this work, we have addressed the fundamentals of using class information for penalty representation regularization. The results indicate that class-wise representation regularizers are very efficient and quite effective, and they should be considered as important and high-potential configurations for learning of deep networks.

\section*{Acknowledgments}
This work was supported by the National Research Foundation of Korea (NRF) grant funded by the Korea government (MSIT) (No. NRF-2017R1E1A1A03070560) and by SK telecom Co., Ltd.

\bibliography{main}
\bibliographystyle{aaai}

\clearpage  
\input{sup_materials.tex} 

\end{document}

%% file: sup_materials.tex
\onecolumn

\begin{center}
	\textbf{\LARGE Supplementary Materials}
\end{center}

\bigskip

\section*{A\quad Architectures and Hyperparameters}

\bigskip

\subsection*{A.1\quad Default Settings}
By default, we chose ReLU, SGD with Adam optimizer, and a learning rate of 0.0001 for networks. Mini-batch size is set to 100 by default but is set to 500 only for CIFAR-100. We evaluated validation performance for \{0.001, 0.01, 0.1, 1, 10, 100\} and chose the one 
with the best performance for each regularizer and condition.
Then, performance was evaluated through five trainings 
using the pre-fixed weight value. In the case of CIFAR-10 and CIFAR-100, 
the last 10,000 instances of 50,000 training data were used as the validation data,
and after the weight values are fixed, the validation data was merged back into training data. All experiments in this work were carried out using TensorFlow 1.5.

\bigskip

\subsection*{A.2\quad MNIST}
For classification tasks, a 6-layer MLP that has 100 hidden units per layer was used. For image reconstruction task, a 6-layer autoencoder was used. The number of hidden units in each layer is 400, 200, 100, 200, 400, and 784 in the order of hidden layers. 

\bigskip

\subsection*{A.3\quad CIFAR-10 and CIFAR-100}
A CNN with four convolutional layers and one fully connected layer was used for both of CIFAR-10 and CIFAR-100. Detailed architecture hyperparameters are shown in Table 6.

\begin{table}[htbp]
\centering
\captionsetup{labelformat=empty}
\caption{Table 6: Default architecture hyperparameters of CIFAR-10/100 CNN model.}
\resizebox{\textwidth}{!}{%
\begin{tabular}{cccccc}
\hline
Layer                 & \# of filters (or units) & Filter size               & Conv. stride & Pooling size              & Pooling stride \\ \hline
Convolutional layer-1 & 32                & 3 $\times$ 3 & 1             & -                         & -              \\
Convolutional layer-2 & 64                & 3 $\times$ 3 & 1             & -                         & -              \\
Max-pooling layer-1   & -                 & -                         & -             & 2 $\times$ 2 & 2              \\
Convolutional layer-3 & 128               & 3 $\times$ 3 & 1             & -                         & -              \\
Max-pooling layer-2   & -                 & -                         & -             & 2 $\times$ 2 & 2              \\
Convolutional layer-4 & 128               & 3 $\times$ 3 & 1             & -                         & -              \\
Max-pooling layer-3   & -                 & -                         & -             & 2 $\times$ 2 & 2 \\ 
Fully connected layer   & 128                 & -                         & -             & - & - \\ \hline             
\end{tabular}
}
\label{table:hyperparameters}
\end{table}

\clearpage

\section*{B\quad Result Details}
\begin{table*}[ht]
\captionsetup{labelformat=empty}
\caption{Table 7: Results for MNIST MLP model. 
The best performing regularizer in each condition (each column) is shown in bold.
For the default condition, the standard values of data size=50k and layer width=100 were used.}
\vskip -0.8in
\begin{center}
\begin{small}
\begin{tabular}{lcccccr}
\hline
\multirow{2}{*}{Regularizer} & \multirow{2}{*}{Default} & \multicolumn{2}{c}{Data size}      & \multicolumn{2}{c}{Layer width}     \\ \cmidrule{3-6} 
                             &                          & 1k               & 5k              & 2                & 8                \\ \hline
Baseline                     & $2.85 \pm 0.11$          & $11.41 \pm 0.19$ & $6.00 \pm 0.07$ & $31.62 \pm 0.07$ & $10.52 \pm 0.57$ \\ \hline
L1W                          & $2.85 \pm 0.06$          & $11.64 \pm 0.27$ & $5.96 \pm 0.11$ & $31.67 \pm 0.15$ & $11.02 \pm 0.58$ \\ 
L2W                          & $3.02 \pm 0.40$          & $11.38 \pm 0.18$ & $5.86 \pm 0.10$ & $31.66 \pm 0.13$ & $10.65 \pm 0.23$ \\ \hline
CR (DeCov)                           & $2.50 \pm 0.05$          & $11.63 \pm 0.24$ & $6.05 \pm 0.06$ & $34.80 \pm 0.25$ & $10.25 \pm 0.74$ \\ 
cw-CR                        & $2.49 \pm 0.10$          & $10.62 \pm 0.05$ & \pmb{$5.80 \pm 0.15$} & $31.50 \pm 0.11$ & $10.81 \pm 1.11$ \\ 
VR                           & $2.65 \pm 0.11$          & $14.42 \pm 0.14$ & $6.90 \pm 0.22$ & $32.39 \pm 0.13$ & \pmb{$9.22 \pm 0.28$}  \\ 
cw-VR                        & \pmb{$2.42 \pm 0.06$}    & \pmb{$10.44 \pm 0.18$} & $5.90 \pm 0.12$ & \pmb{$30.34 \pm 0.06$} & $10.01 \pm 0.63$ \\ 
\hline
\end{tabular}
\label{appendix_mnist}
\end{small}
\end{center}
\vskip 0.1in
\bigskip

\centering
\captionsetup{labelformat=empty}
\caption{Table 8: Results for CIFAR-10 CNN model. 
The best performing regularizer in each condition (each column) is shown in bold.
For the default condition, the standard values of data size=50k and layer width=128 were used 
and Adam optimizer was applied.}
\vskip -0.8in
\begin{center}
\resizebox{\textwidth}{!}{%
\begin{tabular}{cccccccc}
\hline
\multirow{2}{*}{Regularizer} & \multirow{2}{*}{Default} & \multicolumn{2}{c}{Data size}       & \multicolumn{2}{c}{Layer width} & \multicolumn{2}{c}{Optimizer}    \\ \cmidrule{3-8}
                             &                          & 1k               & 5k               & 32                & 512         & {Momentum} & {RMSProp}       \\ \midrule
Baseline                     & $26.64 \pm 0.16$         & $56.07 \pm 0.36$ & $43.95 \pm 0.43$ & $28.54 \pm 0.63$ & $28.52 \pm 1.06$             & $25.78 \pm 0.37$ & $28.52 \pm 1.21$ \\ \hline
L1W                          & $26.46 \pm 0.39$         & $56.64 \pm 0.91$ & $44.32 \pm 0.66$ & $28.65 \pm 1.14$ & $27.96 \pm 0.72$             & $25.73 \pm 0.40$ & $28.30 \pm 0.99$ \\
L2W                          & $25.71 \pm 0.98$            & $56.57 \pm 0.22$ & $44.87 \pm 0.81$ & $28.54 \pm 0.30$  & $27.79 \pm 0.83$    & $26.35 \pm 0.54$ & $28.02 \pm 0.88$ \\ \hline
CR (DeCov)                          & $24.96 \pm 0.63$         & $57.40 \pm 2.11$  & $45.16 \pm 0.94$ & $26.45 \pm 0.22$ & $28.65 \pm 1.21$            & $26.72 \pm 0.61$   & $27.94 \pm 0.43$  \\
cw-CR                        & $22.99 \pm 0.58$         & $53.50 \pm 1.05$  & \pmb{$42.15 \pm 0.64$} & $26.40 \pm 0.62$  & $28.54 \pm 1.01$    & $25.93 \pm 0.59$    & $27.77 \pm 0.88$  \\
VR                           & \pmb{$21.44 \pm 0.88$}         & $53.90 \pm 0.97$  & $42.33 \pm 0.57$ & \pmb{$24.96 \pm 0.26$} & $26.61 \pm 0.47$ & $25.01 \pm 0.41$    & \pmb{$26.06 \pm 0.72$}  \\
cw-VR                        & $21.58 \pm 0.21$         & \pmb{$51.93 \pm 1.09$} & $43.00 \pm 0.95$    & $25.81 \pm 0.64$ & \pmb{$26.46 \pm 0.25$}     &  \pmb{$24.42 \pm 0.31$}   & $26.19 \pm 1.35$  \\
\hline
\end{tabular}%
}
\label{cifar10_dependency}
\end{center}
\vskip 0.1in

\bigskip


\centering
    \captionsetup{labelformat=empty}
\caption{Table 9: Results for CIFAR-100 CNN model. The best performing regularizer in each condition (each column) is shown in bold. For the default condition, the standard values of data size=50k, layer width=128, and number of classes=100 were used.}
\vskip -0.8in
\begin{center}
\resizebox{\textwidth}{!}{%
\begin{tabular}{ccccccccc}
\hline
\multirow{2}{*}{Regularizer} & \multirow{2}{*}{Default} & \multicolumn{2}{c}{Data Size}       & \multicolumn{2}{c}{Layer Width}     & \multicolumn{3}{c}{Classes}                            \\ \cmidrule{3-9}
                             &                          & 1k               & 5k               & 32               & 512              & 4                & 16               & 64               \\ \midrule
Baseline                     & $61.26 \pm 0.52$         & $90.89 \pm 0.30$ & $82.21 \pm 0.72$ & $62.41 \pm 0.34$ & $61.30 \pm 0.64$  & \pmb{$24.95 \pm 2.36$} & $45.75 \pm 0.73$ & $58.02 \pm 0.40$  \\ \hline
L1W                          & $60.97 \pm 0.64$         & $91.33 \pm 0.37$ & $82.3 \pm 0.6$   & $62.23 \pm 0.58$ & $60.92 \pm 0.47$ & $26.75 \pm 2.04$ & $45.08 \pm 1.53$ & $58.08 \pm 1.18$ \\
L2W                          & $60.23 \pm 0.31$         & $90.53 \pm 0.39$ & $82.05 \pm 0.70$  & $62.78 \pm 0.36$ & $61.55 \pm 0.99$ & $26.90 \pm 1.24$  & $45.28 \pm 1.59$ & $57.47 \pm 0.66$ \\ \hline
CR (DeCov)                          & $59.88 \pm 0.50$         & $91.70 \pm 0.14$  & $82.47 \pm 0.41$ & \pmb{$60.47 \pm 0.63$} & $60.70 \pm 0.94$  & $27.25 \pm 1.51$ & $44.55 \pm 1.10$  & $56.76 \pm 0.86$ \\
cw-CR                        & $57.03 \pm 0.73$         & $90.85 \pm 0.29$ & $81.29 \pm 0.62$ & $61.41 \pm 0.67$ & $58.02 \pm 0.25$ & $26.35 \pm 1.04$ & $43.50 \pm 1.21$  & $54.24 \pm 0.64$ \\ 
VR                           & $57.68 \pm 0.94$         & $91.43 \pm 0.32$ & $81.85 \pm 0.38$ & $61.35 \pm 0.45$ & \pmb{$56.87 \pm 0.74$} & $26.10 \pm 1.81$  & $42.33 \pm 1.03$ & $54.32 \pm 0.40$  \\
cw-VR                        & \pmb{$56.75 \pm 0.64$}   & \pmb{$90.45 \pm 0.22$} & \pmb{$81.03 \pm 0.57$} & $60.67 \pm 0.59$ & $56.91 \pm 0.73$ & $26.40 \pm 1.08$  & \pmb{$41.38 \pm 0.53$} & \pmb{$54.23 \pm 1.06$} \\ \hline

\end{tabular}%
}
\label{cifar100_dependency}
\end{center}
\vskip 0.1in

\bigskip


\centering
\captionsetup{labelformat=empty}
\caption{Table 10: Mean squared error of deep autoencoder.}
\begin{tabular}{cc}
\hline
Regularizer & Mean Squared Error                       \\ \hline
Baseline    & $1.44 \times 10^{-2} \pm 3.36 \times 10^{-4} $                        \\ \hline
CR          & $1.29 \times 10^{-2} \pm 2.44 \times 10^{-4} $                        \\ 
cw-CR       & $1.22 \times 10^{-2} \pm 3.63 \times 10^{-4} $                        \\ 
VR          & $1.29 \times 10^{-2} \pm 5.16 \times 10^{-4} $                        \\
cw-VR       & \pmb{$1.19 \times 10^{-2} \pm 2.48 \times 10^{-4} $}                  \\ \hline
\end{tabular}
\label{table:autoencoder}
\vskip -1.2in
\end{table*}

\clearpage

\section*{C\quad Principal Component Analysis of Learned Representations}

\begin{figure}[htbp]
    \centering
    \quad\subfloat[Baseline (Before ReLU)]{{\includegraphics[width=4.1cm]{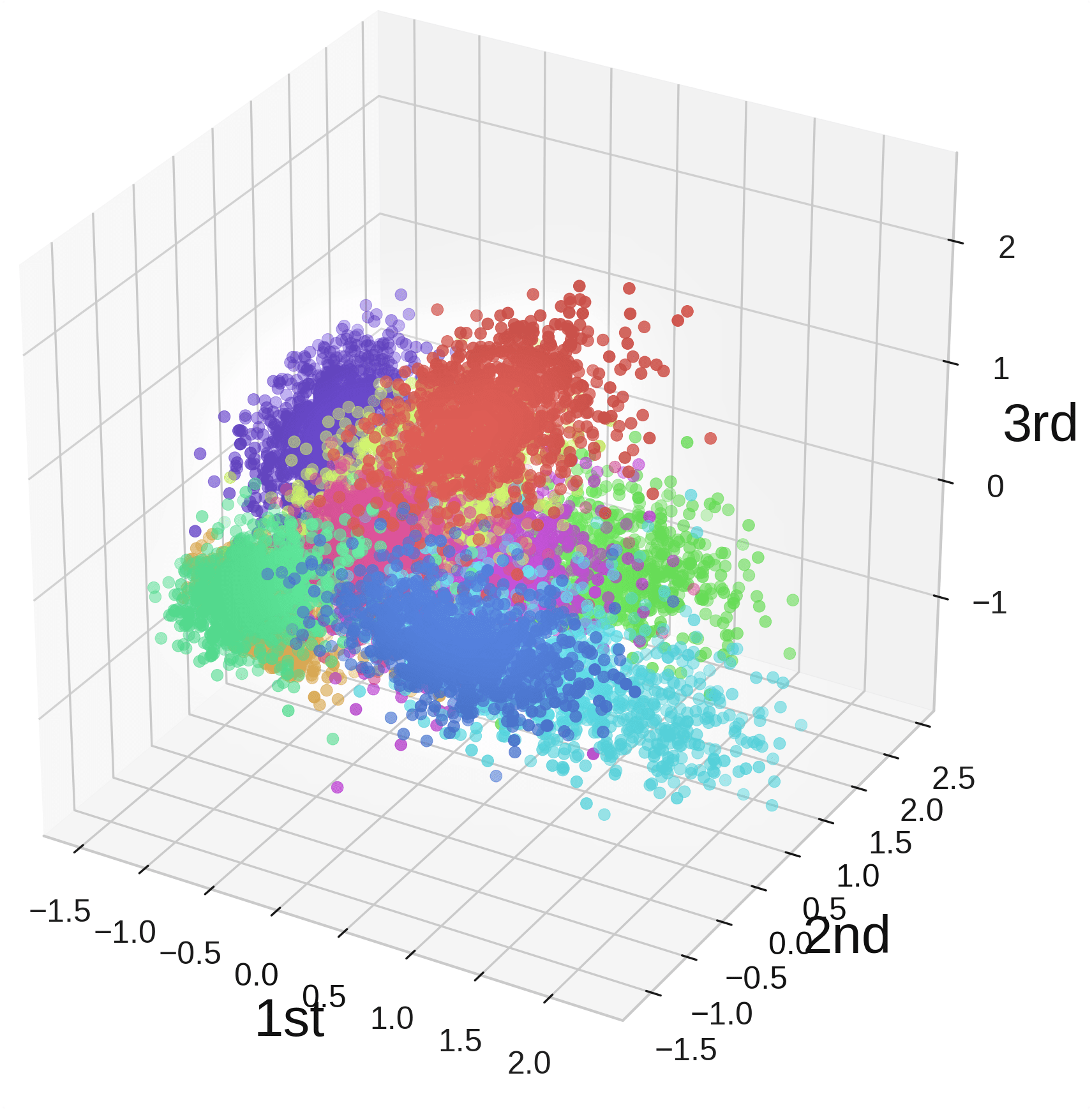} }}%
    \qquad\qquad\qquad\quad
    \subfloat[Baseline (After ReLU)]{{\includegraphics[width=4.1cm]{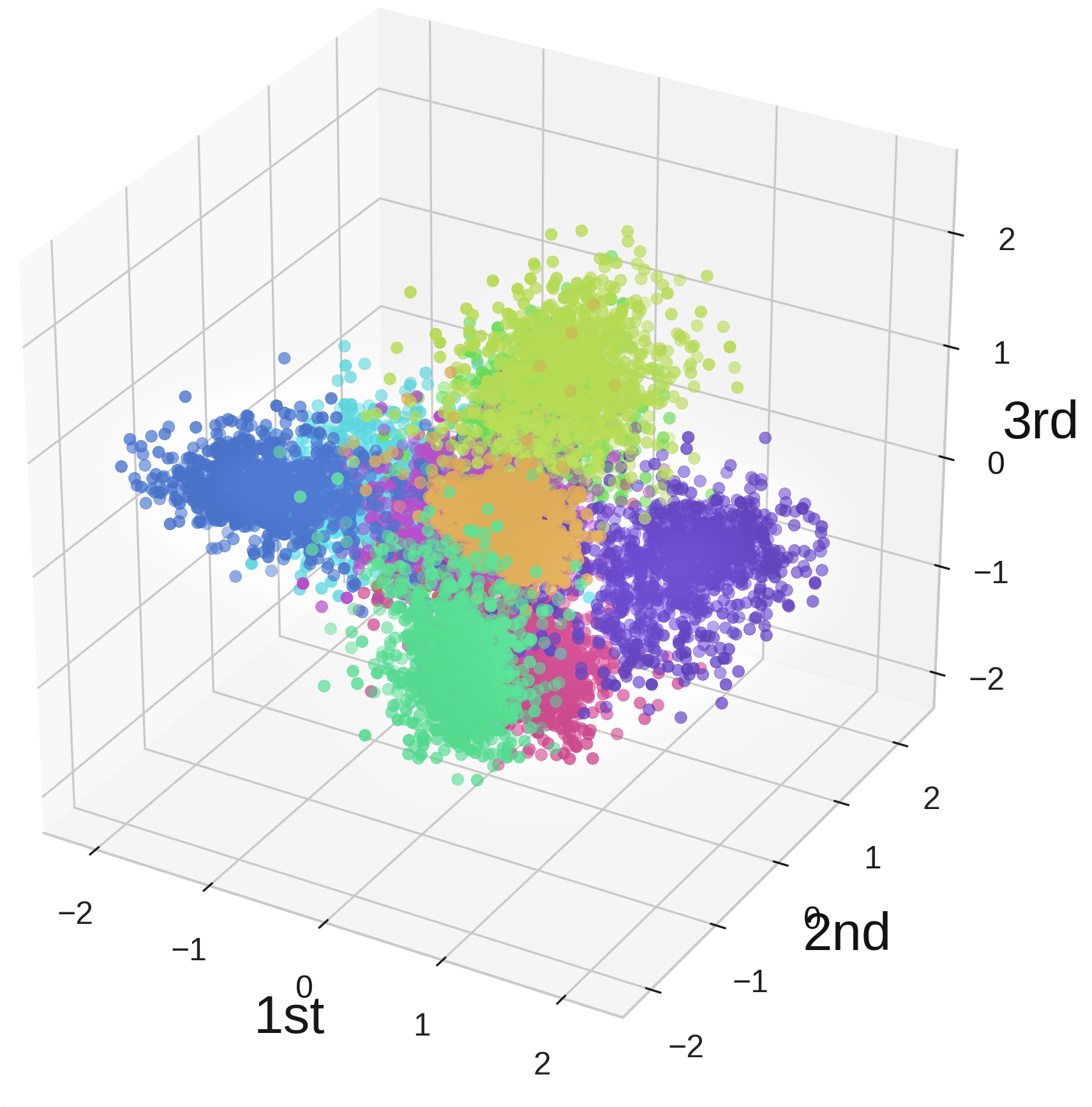} }} 
    
    \subfloat[L1W (Before ReLU)]{{\includegraphics[width=4.5cm]{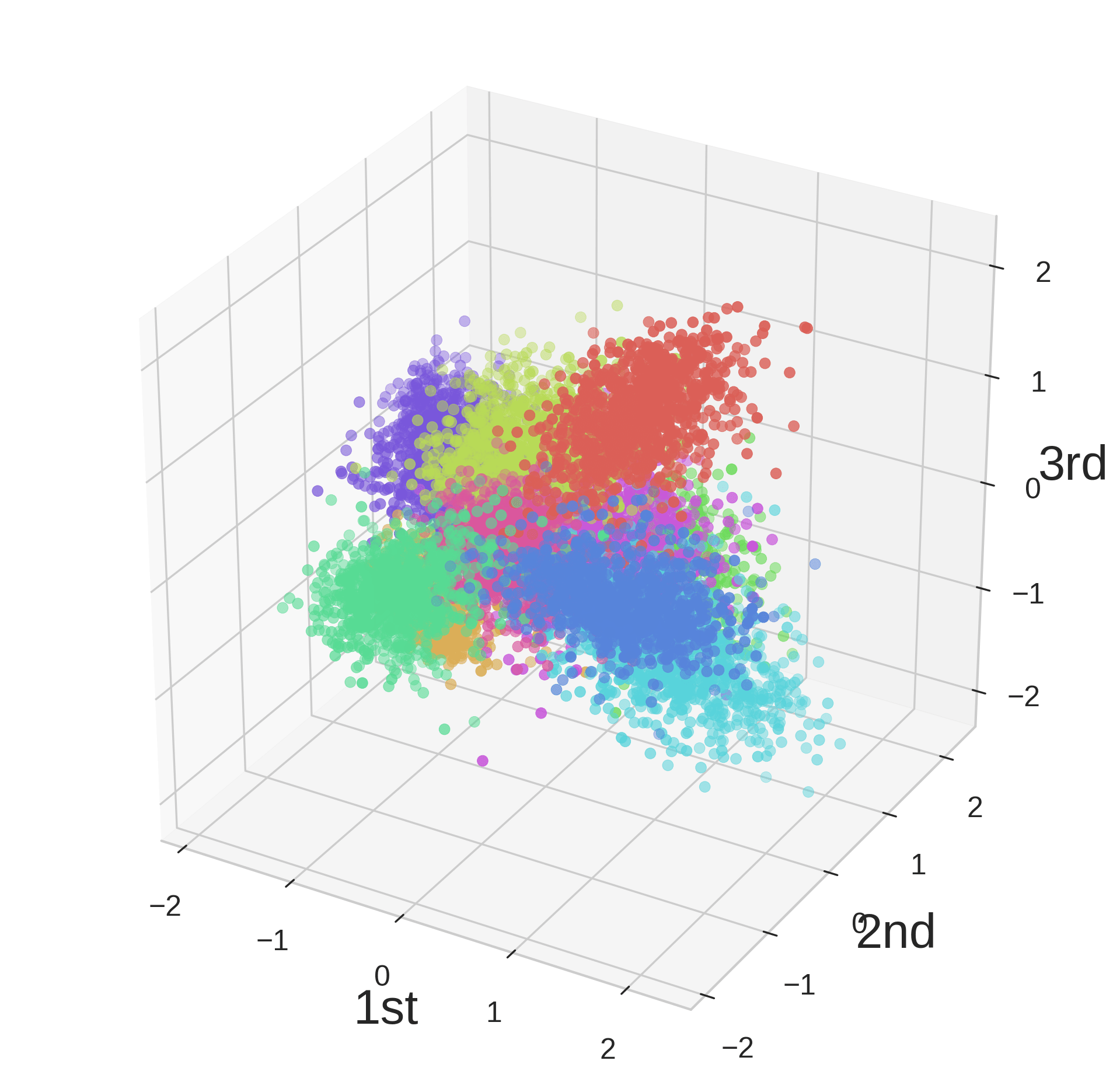} }}%
    \qquad\qquad\qquad
    \subfloat[L1W (After ReLU)]{{\includegraphics[width=4.5cm]{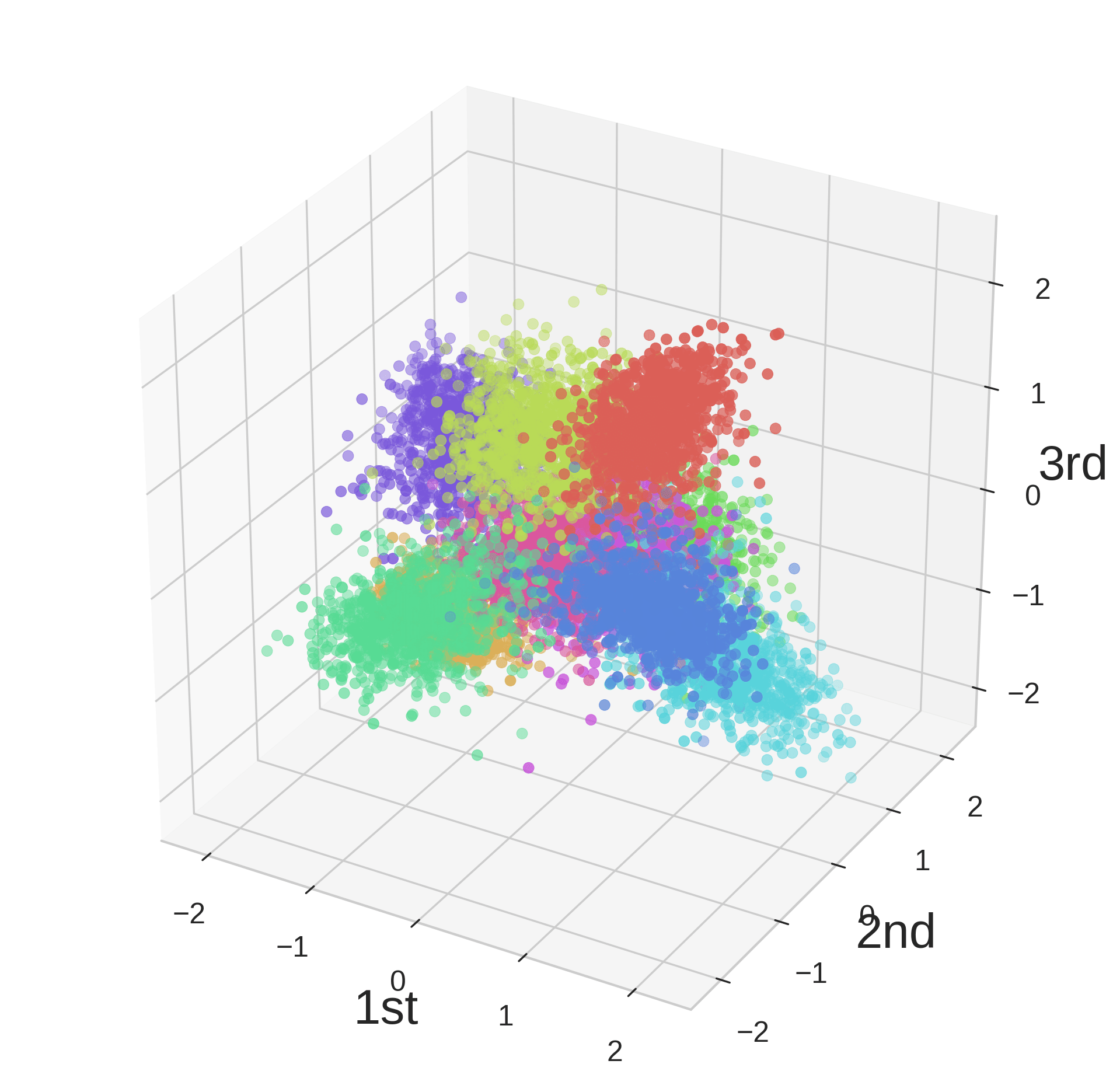} }} 
    
    \subfloat[L2W (Before ReLU)]{{\includegraphics[width=4.5cm]{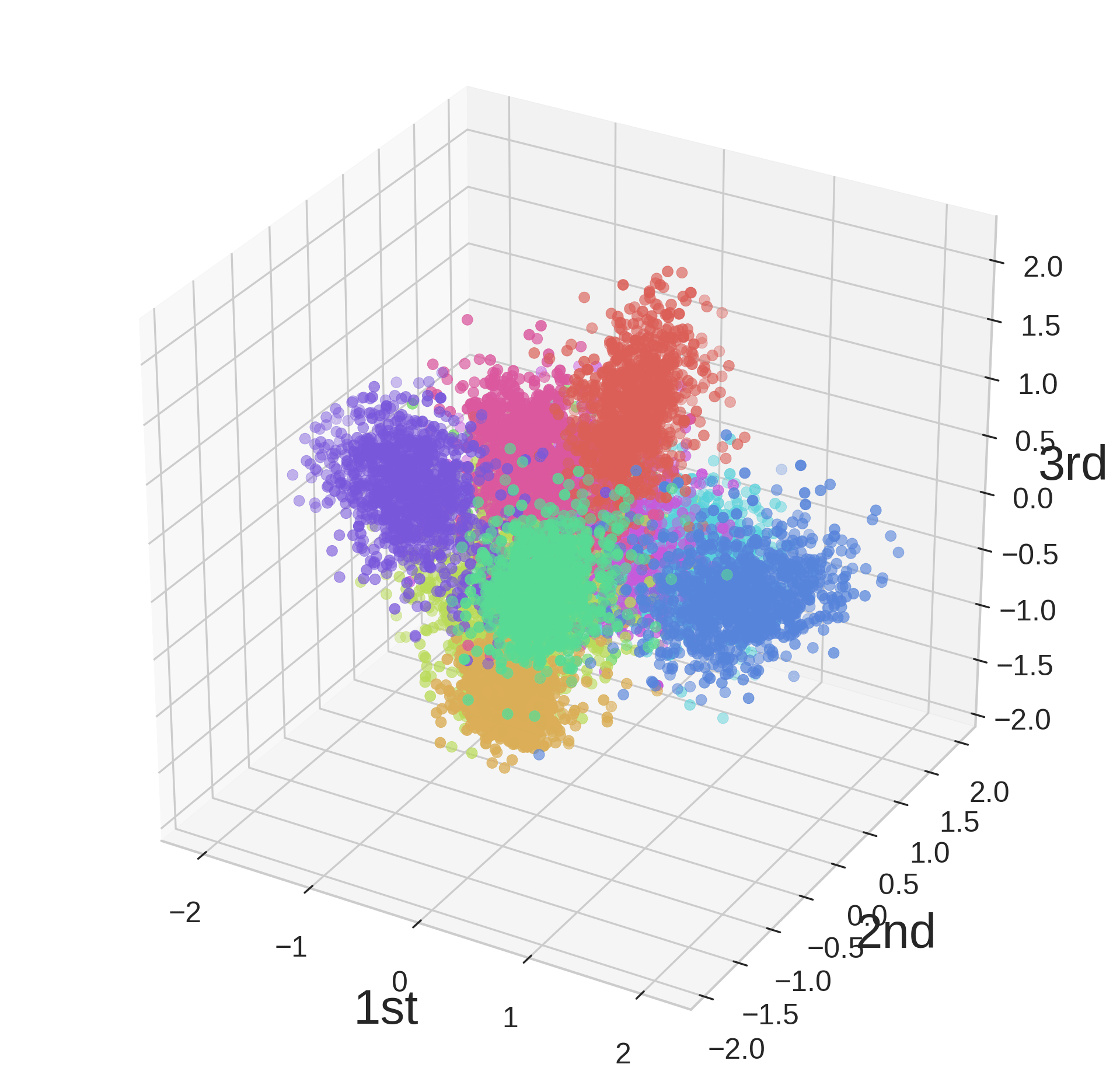} }}%
    \qquad\qquad\qquad
    \subfloat[L2W (After ReLU)]{{\includegraphics[width=4.5cm]{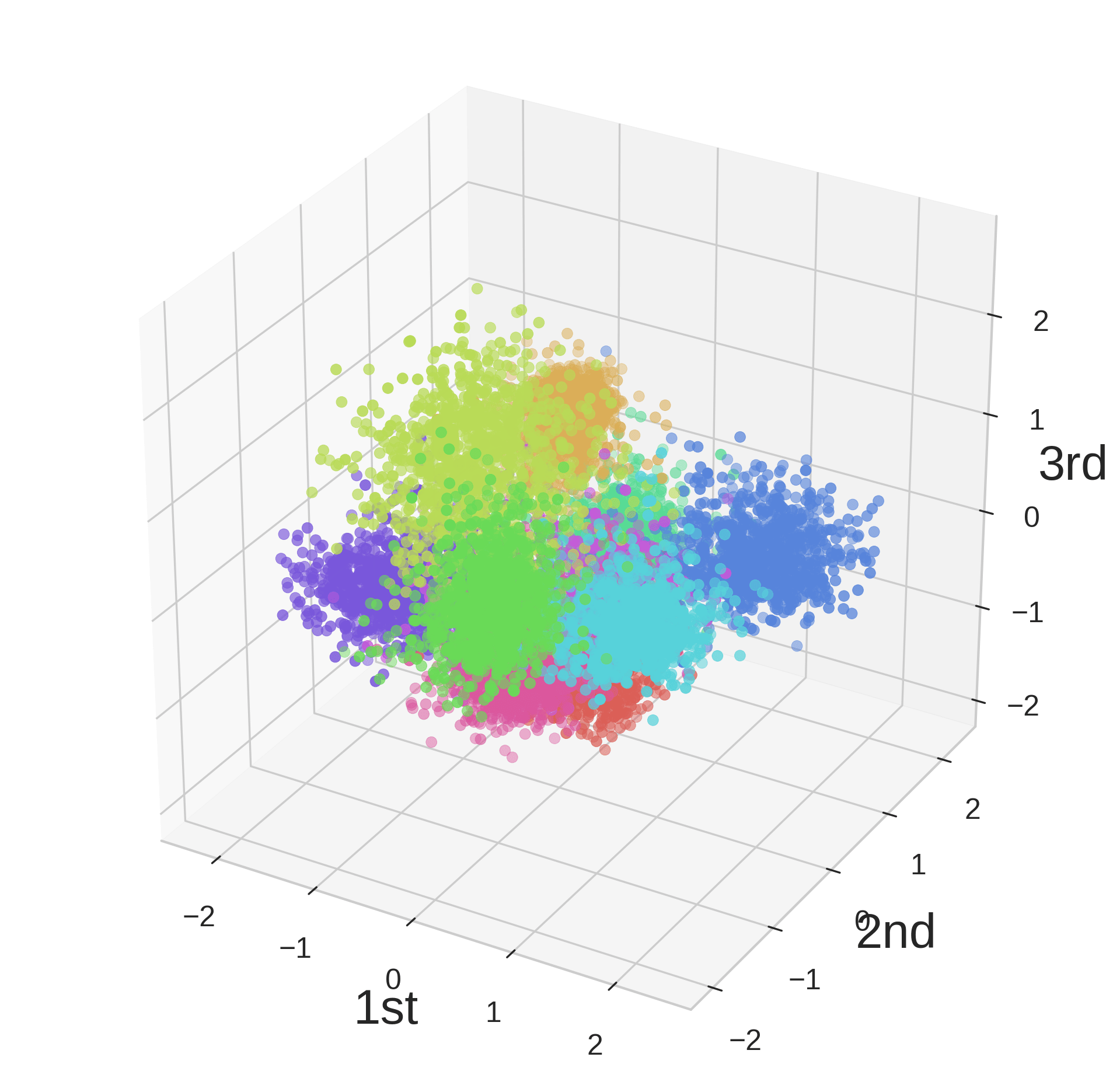} }} 
     
    \captionsetup{labelformat=empty}
\caption{Figure 4: The top three principal components of learned representations (Baseline, L1W, and L2W). Note that representation characteristics of L1W and L2W are very similar to those of the baseline because weight decay methods do not directly shape representations.}%
    \label{fig:pca_1}%
\end{figure}

\begin{figure}[htbp]
    \centering
    \subfloat[CR (Before ReLU)]{{\includegraphics[width=4.5cm]{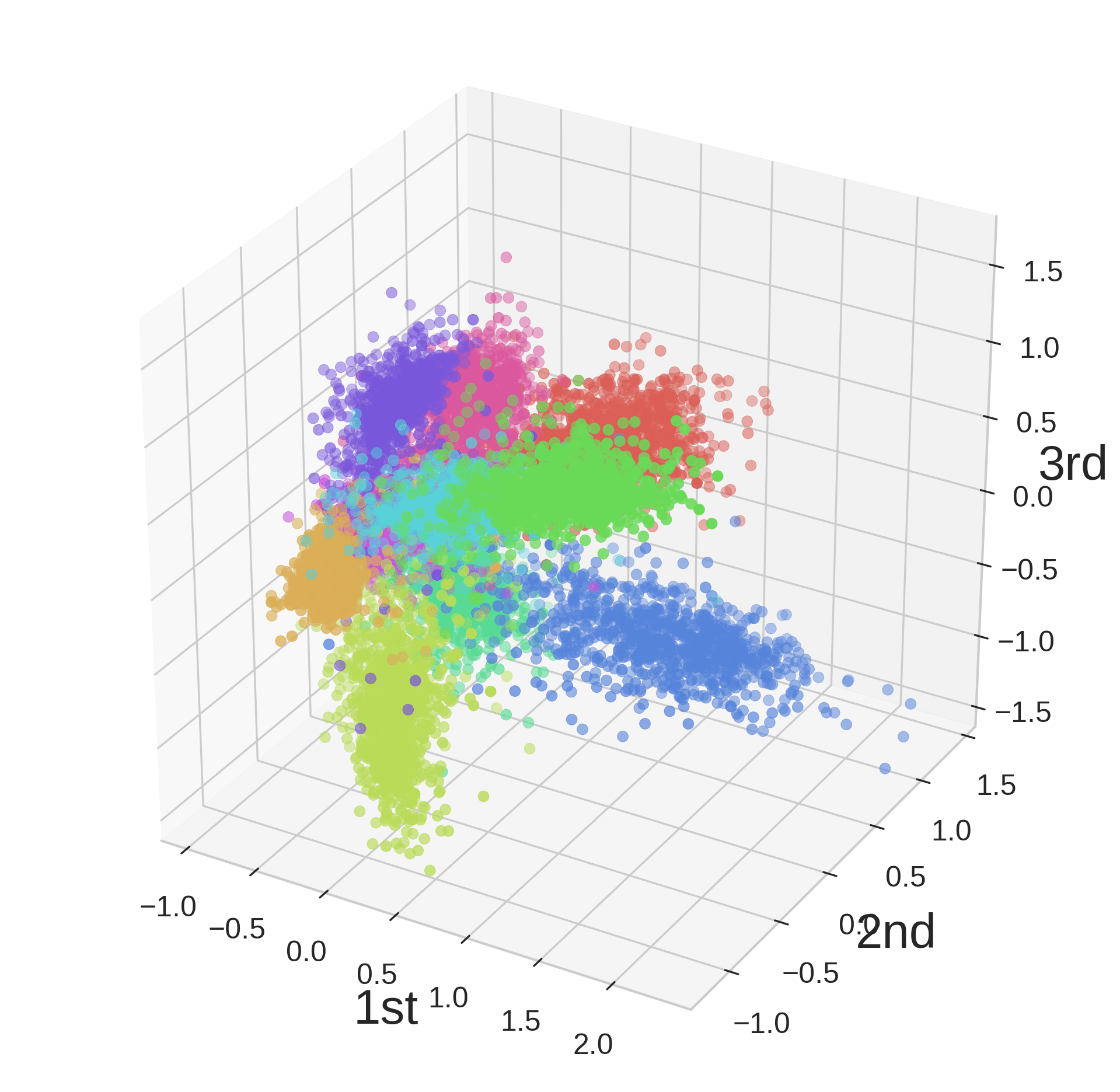} }}%
    \qquad\qquad\qquad
    \subfloat[CR (After ReLU)]{{\includegraphics[width=4.5cm]{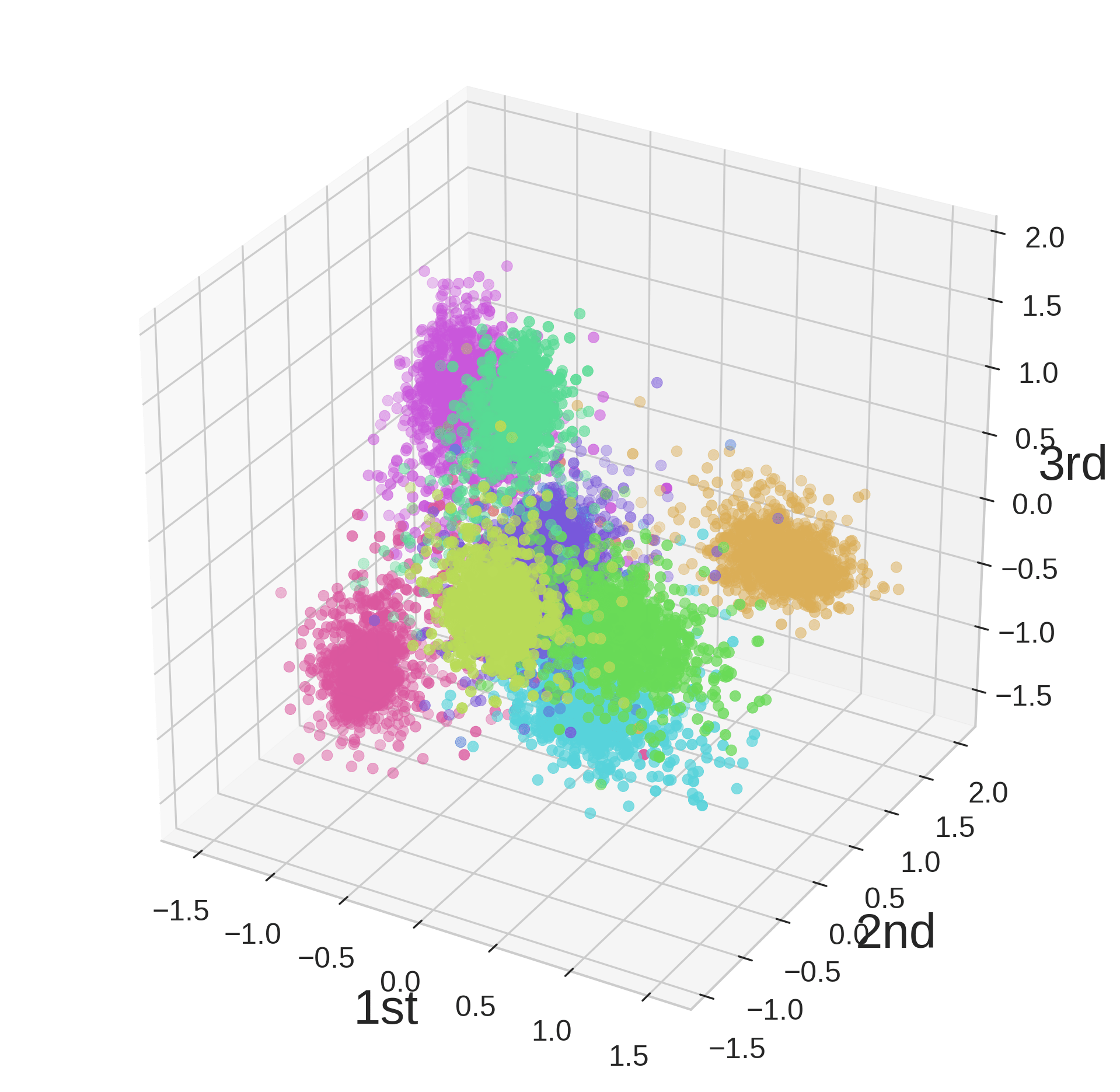} }}
    
    \subfloat[cw-CR (Before ReLU)]{{\includegraphics[width=4.5cm]{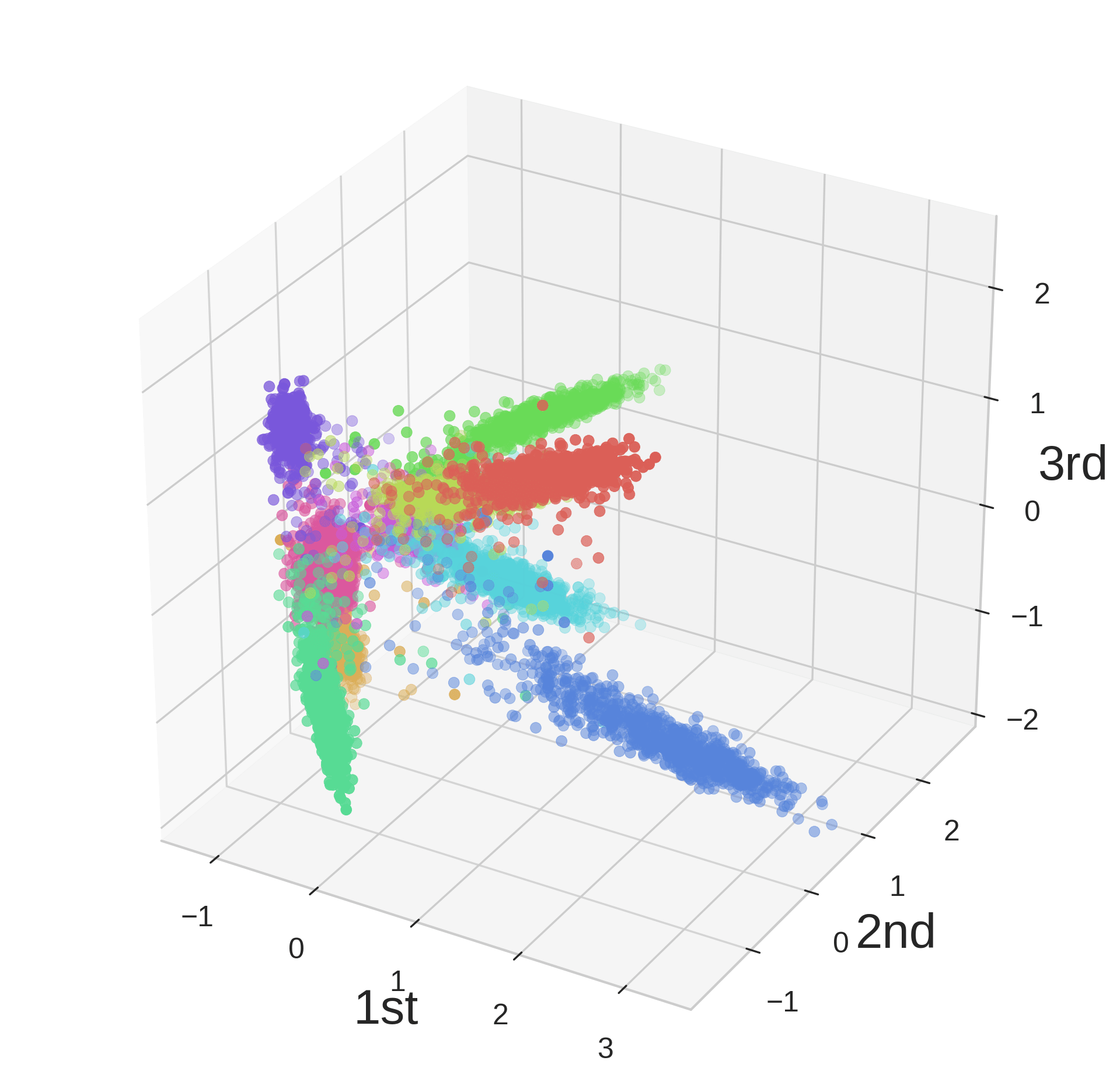} }}%
    \qquad\qquad\qquad
    \subfloat[cw-CR (After ReLU)]{{\includegraphics[width=4.5cm]{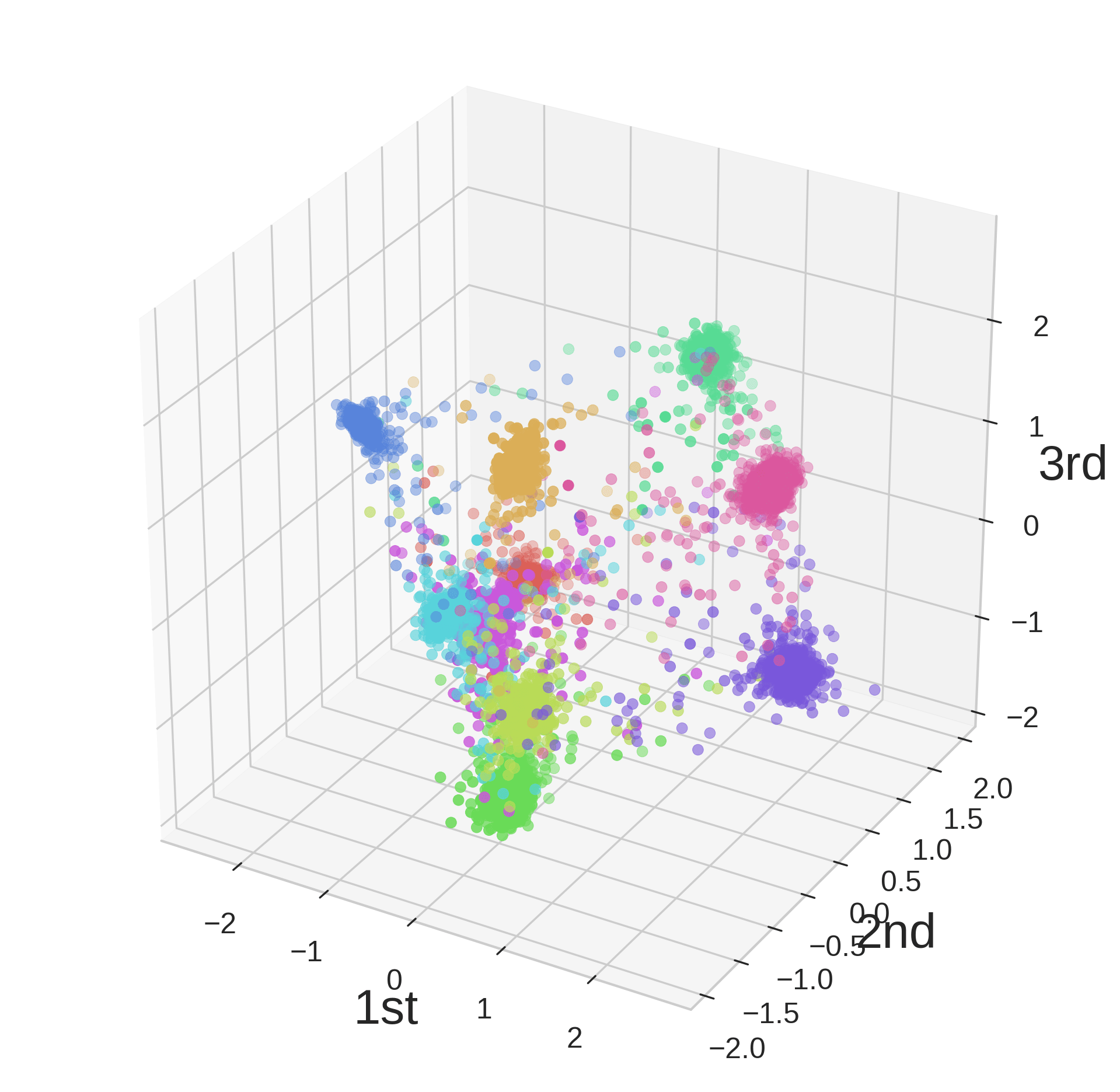} }}
    
    \subfloat[VR (Before ReLU)]{{\includegraphics[width=4.5cm]{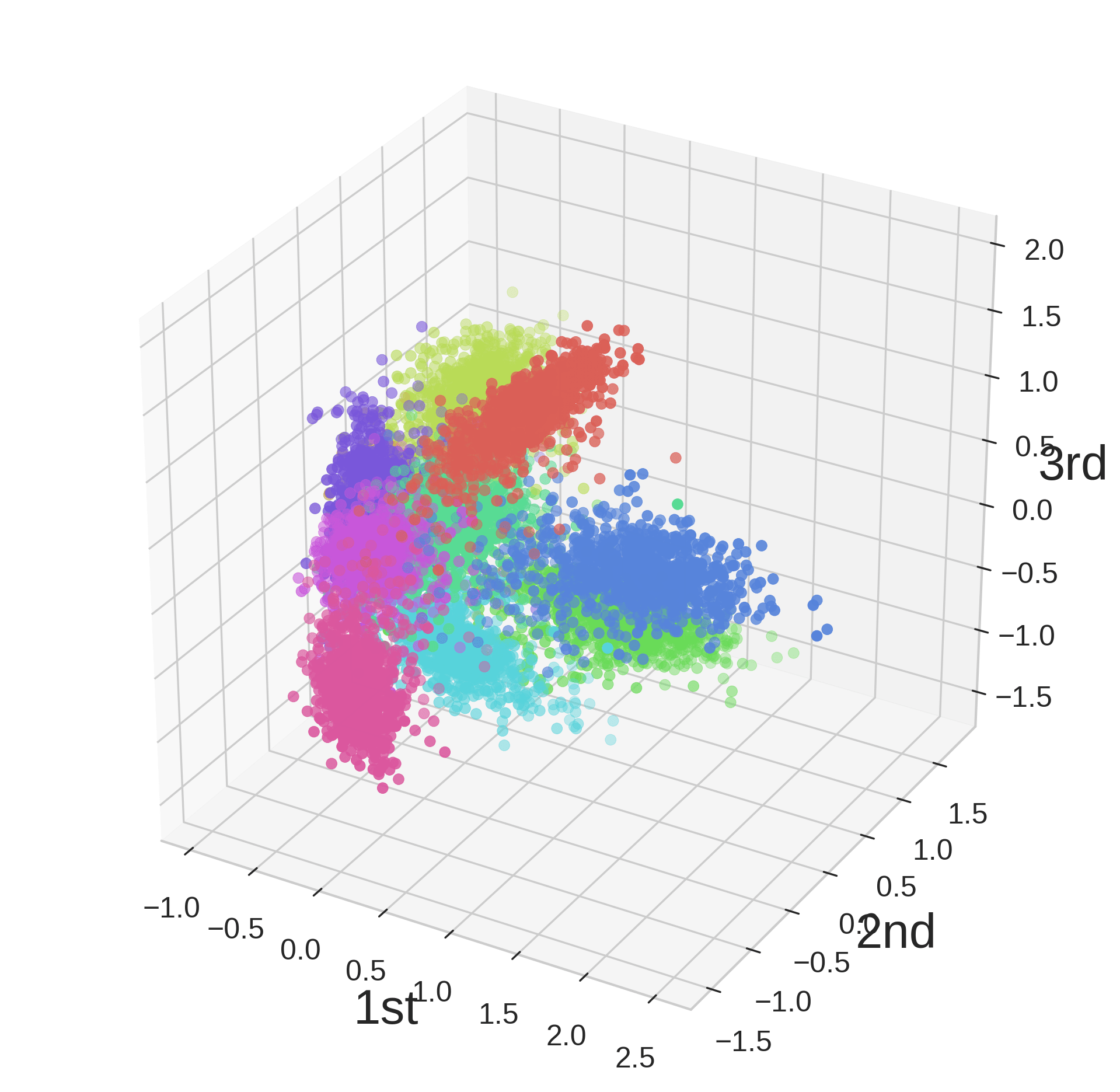} }}%
    \qquad\qquad\qquad
    \subfloat[VR (After ReLU)]{{\includegraphics[width=4.5cm]{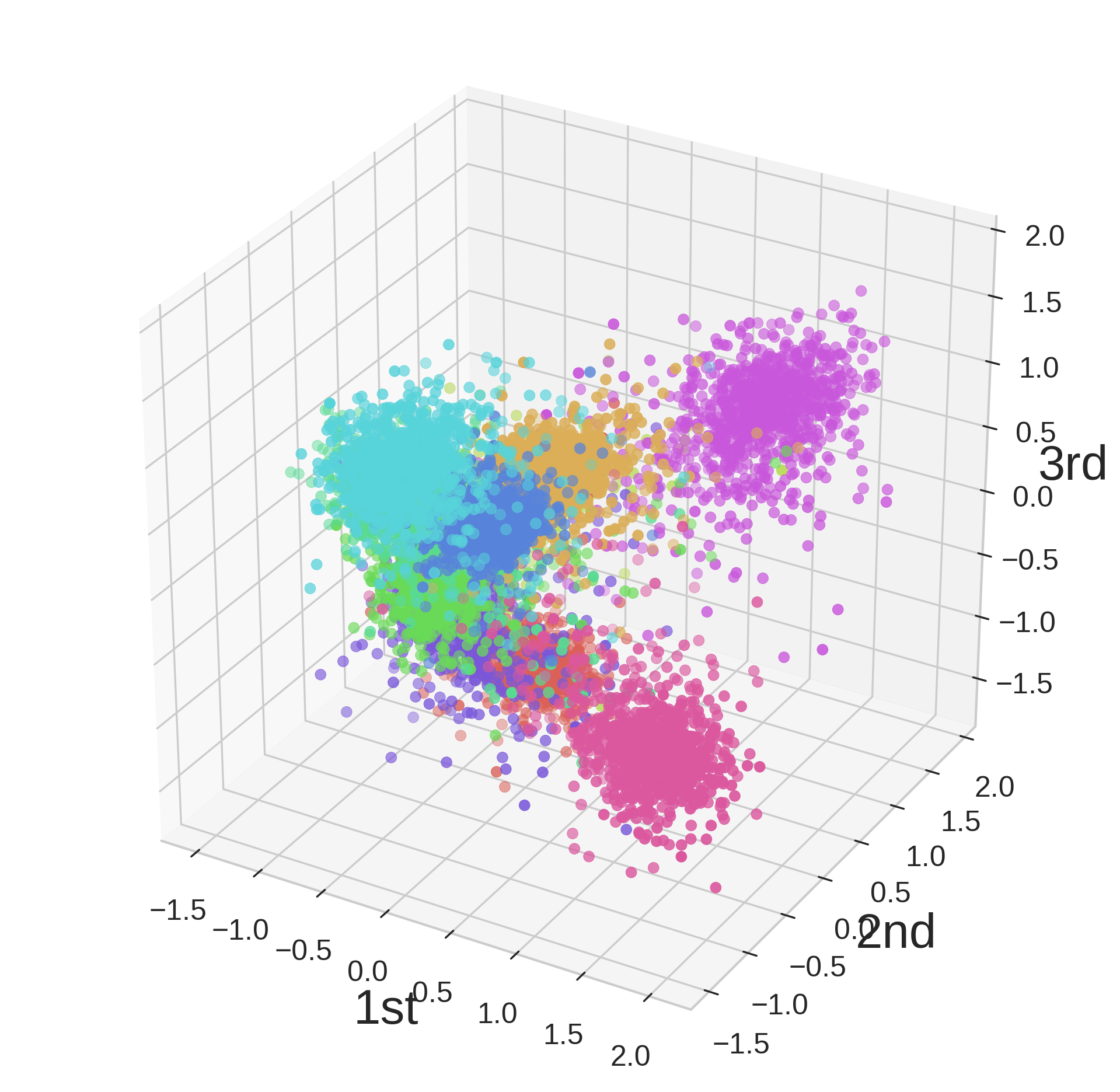} }}  
    
    \qquad\subfloat[cw-VR (Before ReLU)]{{\includegraphics[width=4.cm]{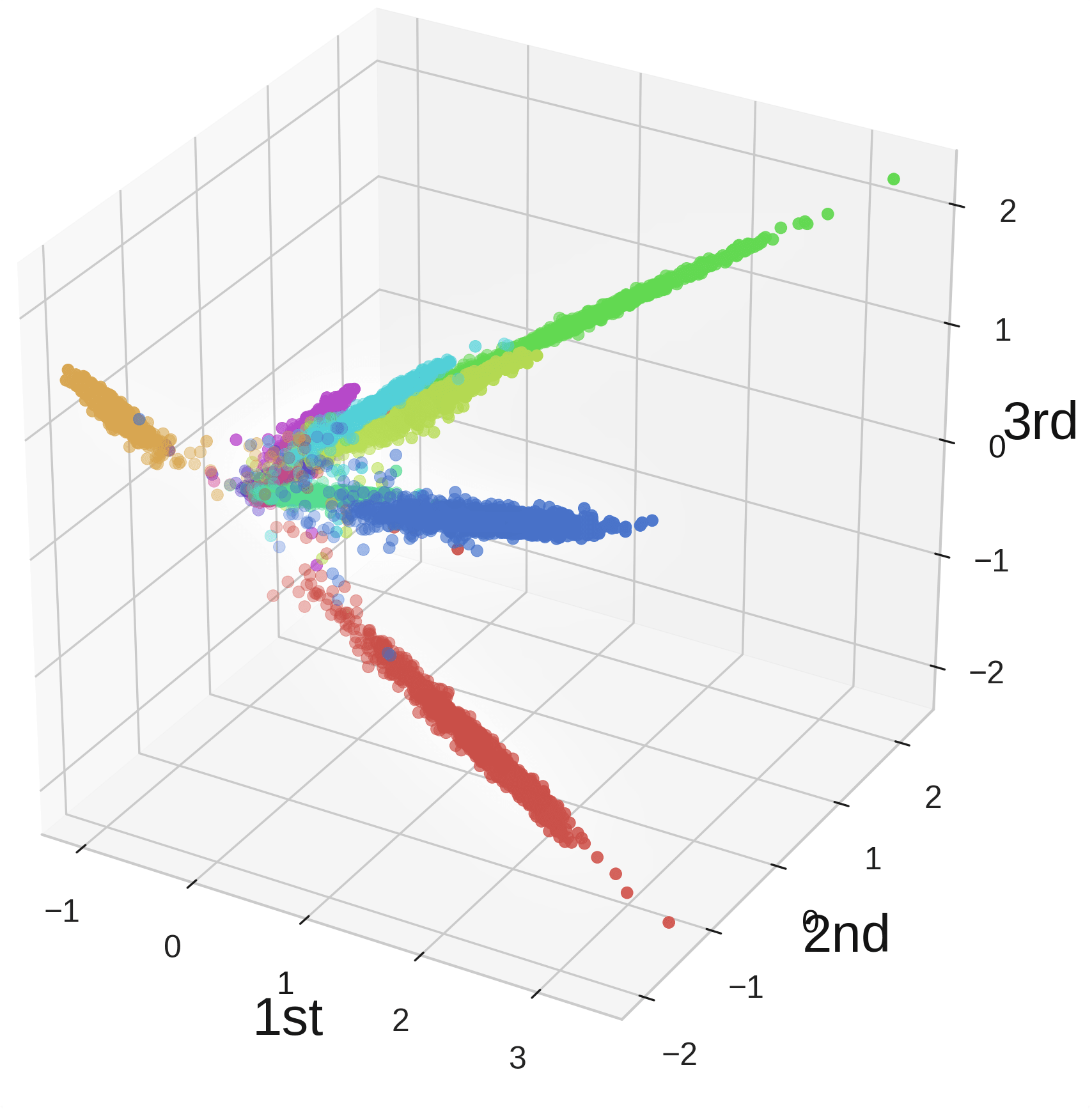} }}%
    \qquad\qquad\qquad\qquad
    \subfloat[cw-VR (After ReLU)]{{\includegraphics[width=4.cm]{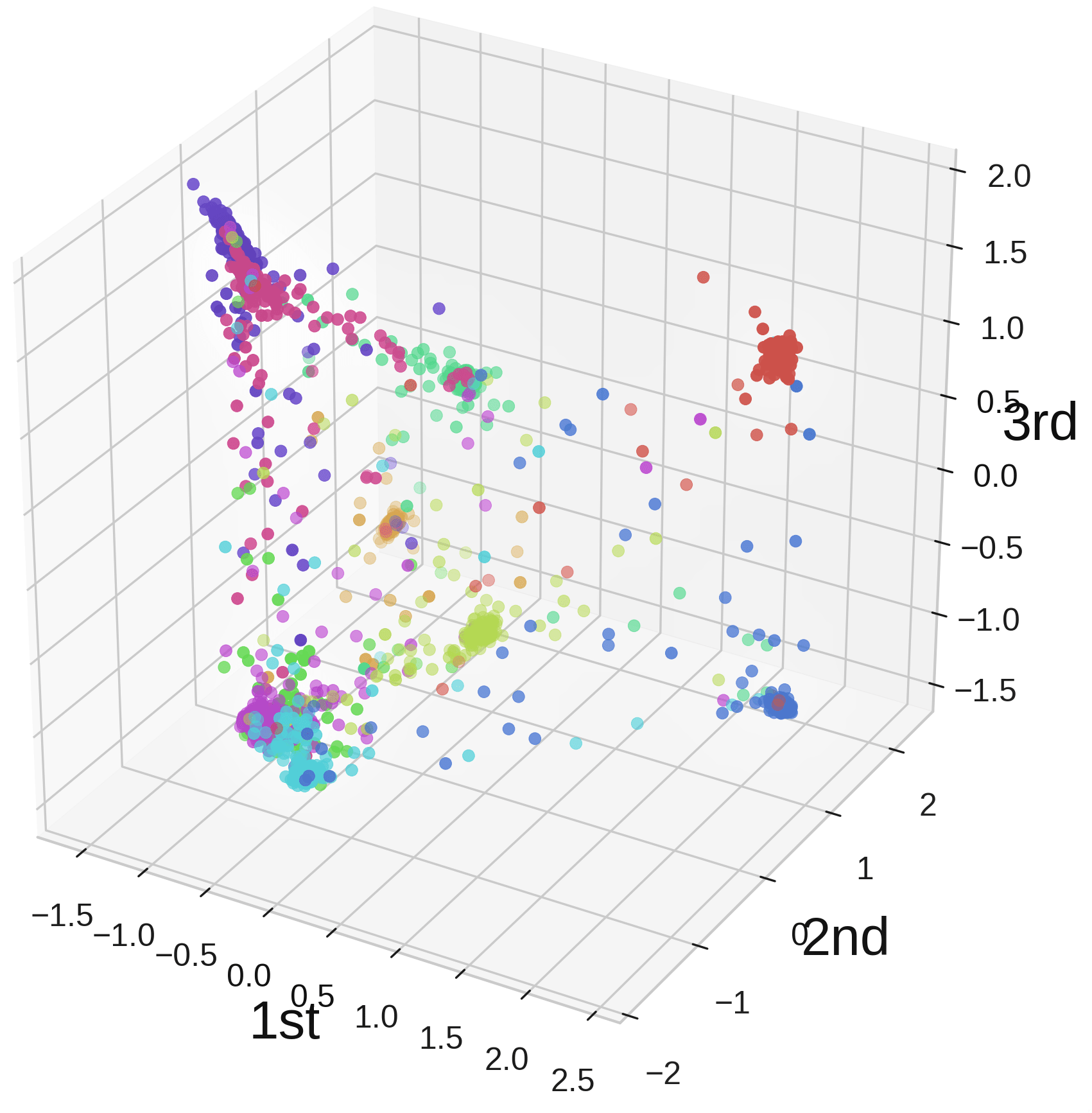} }}  
    \captionsetup{labelformat=empty}
\caption{Figure 5: The top three principal components of learned representations (representation regularizers).}%
    \label{fig:pca_2}%
\end{figure}

\clearpage

\section*{D\quad Layer Dependency}

\begin{figure}[htbp]
\begin{center}
\centerline{\includegraphics[width=4.1cm]{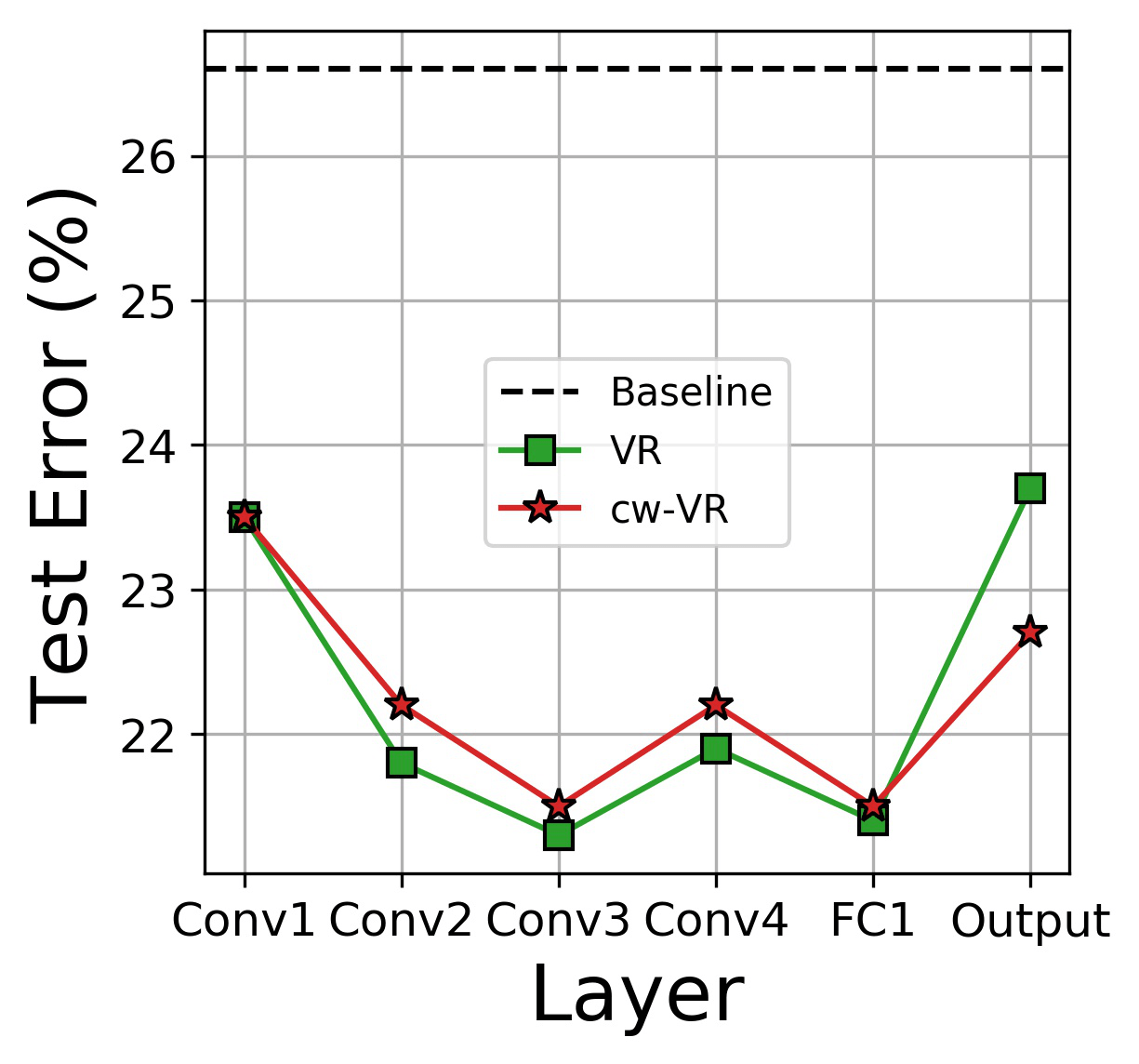}}
\captionsetup{labelformat=empty}
\caption{Figure 6:Layer dependency of representation regularizers on CIFAR-10 CNN model. The x-axis indicates layers where regularizers are applied. CR and cw-CR are excluded because of the high computational burden of applying them to the convolutional layers.}
\label{fig:layer_dependency_cifar10}
\end{center}
\end{figure}